\documentclass[10pt,twocolumn,letterpaper]{article}

\usepackage[pagenumbers]{cvpr} 

\usepackage[dvipsnames]{xcolor}

\newcommand{\parsection}[1]{\noindent\textbf{#1}:}
\usepackage{multirow}
\usepackage{placeins}
\usepackage{siunitx}

\usepackage{xr}
\usepackage[normalem]{ulem}
\usepackage[accsupp]{axessibility}

\definecolor{cvprblue}{rgb}{0.21,0.49,0.74}
\usepackage[pagebackref,breaklinks,colorlinks,citecolor=cvprblue]{hyperref}

\def\paperID{4} 
\def\confName{CVPR}
\def\confYear{2024}

\title{Are NeRFs ready for autonomous driving? \\Towards closing the real-to-simulation gap}

\author{Carl Lindström$^{\dagger,1,2}$
\quad Georg Hess$^{\dagger,1,2}$
\quad Adam Lilja$^{1,2}$
\\ Maryam Fatemi$^{1}$
\quad Lars Hammarstrand$^{2}$
\quad Christoffer Petersson$^{1,2}$
\quad Lennart Svensson$^{2}$
\\
\normalsize$^1$Zenseact \hspace{0.8cm} $^2$Chalmers University of Technology \hspace{0.8cm}\\
{\tt\small \{firstname.lastname\}@\{zenseact.com, chalmers.se\}}
}

\begin{document}
\crefname{subparagraph}{paragraph}{paragraphs}
\Crefname{subparagraph}{Paragraph}{Paragraphs}
\crefname{paragraph}{paragraph}{paragraphs}
\Crefname{paragraph}{Paragraph}{Paragraphs}
\maketitle
\begin{abstract}
Neural Radiance Fields (NeRFs) have emerged as promising tools for advancing autonomous driving (AD) research, offering scalable closed-loop simulation and data augmentation capabilities. 
However, to trust the results achieved in simulation, one needs to ensure that AD systems perceive real and rendered data in the same way. 
Although the performance of rendering methods is increasing, many scenarios will remain inherently challenging to reconstruct faithfully. 
To this end, we propose a novel perspective for addressing the real-to-simulated data gap. 
Rather than solely focusing on improving rendering fidelity, we explore simple yet effective methods to enhance perception model robustness to NeRF artifacts without compromising performance on real data. 
Moreover, we conduct the first large-scale investigation into the real-to-simulated data gap in an AD setting using a state-of-the-art neural rendering technique. 
Specifically, we evaluate object detectors and an online mapping model on real and simulated data, and study the effects of different fine-tuning strategies.
Our results show notable improvements in model robustness to simulated data, even improving real-world performance in some cases. 
Last, we delve into the correlation between the real-to-simulated gap and image reconstruction metrics, identifying FID and LPIPS as strong indicators. 
See \href{https://research.zenseact.com/publications/closing-real2sim-gap}{here} for our project page.
\end{abstract} 

\def\thefootnote{$\dagger$}\footnotetext{These authors contributed equally to this work.}\def\thefootnote{\arabic{footnote}}    
\section{Introduction}
\label{sec:intro}
The development of autonomous vehicles (AVs) requires substantial and accurate testing to ensure safe behavior when deployed in the real world. 
In general, this has required practitioners to collect vast amounts of real-world data. Unfortunately, such collection is time-consuming and limits which safety-critical scenarios can be explored as to not risking the safety of other road users.

\begin{figure}[t]
    \centering
    \includegraphics[width=\linewidth]{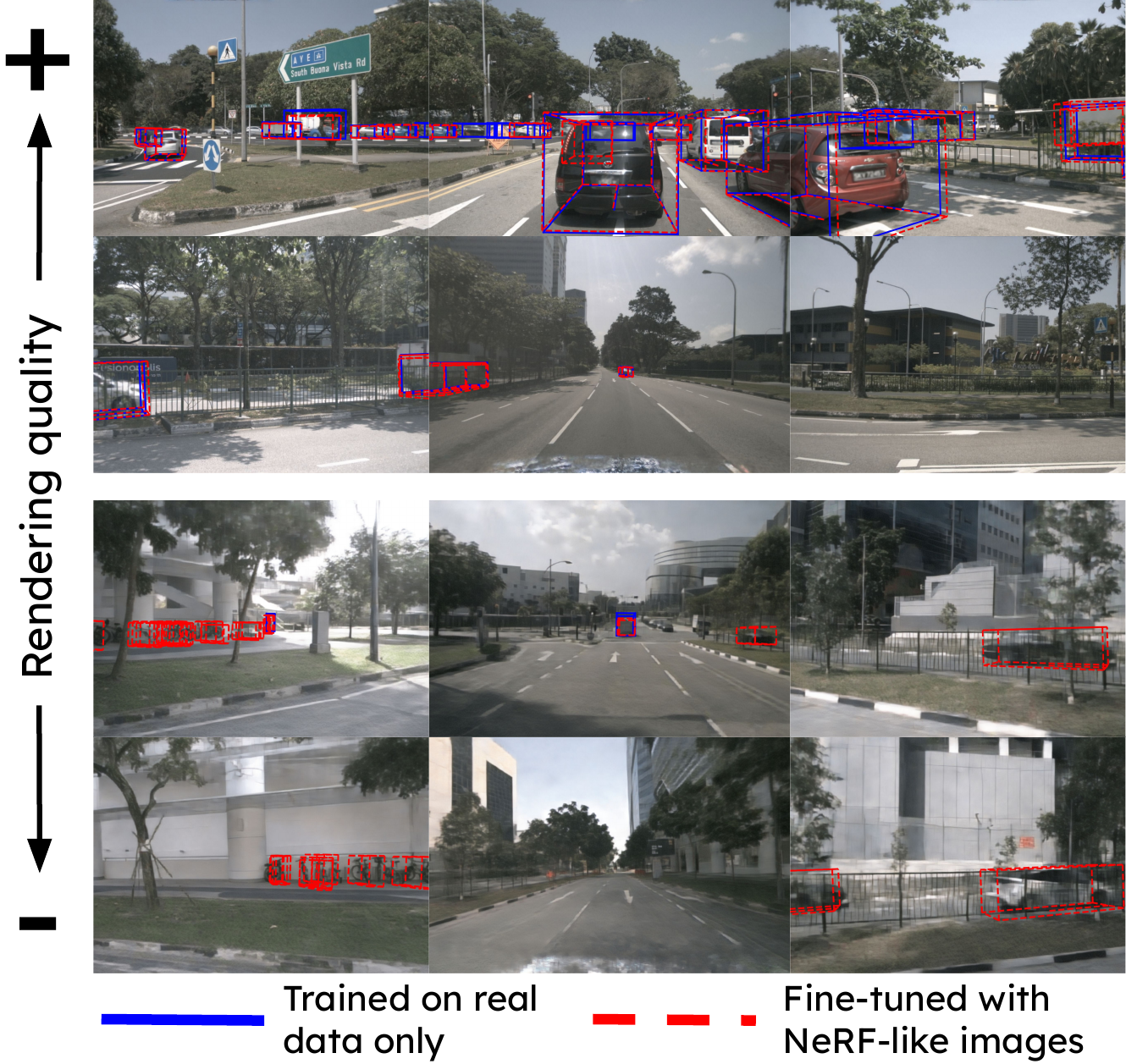}
    \caption{Using NeRFs for autonomous driving testing requires perception models to treat rendered and real images similarly. A BEVFormer model trained on real data detects objects (blue) in high-quality renderings (top). However, when quality decreases (bottom), e.g., scenes challenging for the NeRF, the same model fails to detect even close-by cars. Instead of emphasizing rendering fidelity, we propose to make models robust to these distortions. Fine-tuning the same model on NeRF-like images (red) reduces the real-to-sim gap without harming real-world performance.}
    \label{fig:front-figure}
    \vspace{-18pt}
\end{figure}

Neural rendering techniques, such as Neural Radiance Fields (NeRFs) \cite{mildenhall2020nerf} or Gaussian Splatting \cite{kerbl3Dgaussians}, provide an attractive alternative, as they can be used to simulate new scenarios in already collected data. 
Consequently, they enable practitioners to explore system behavior, from pixel to torque, for safety-critical scenarios that would be difficult to collect in the real world. Multiple works have recently explored how to apply NeRFs to autonomous driving (AD) data \cite{yang2023unisim,tonderski2023neurad,xie2023snerf,wu2023mars,zhou2023drivinggaussian,ljungbergh2024neuroncap}. With constantly increasing rendering quality, and decreasing computational demands, these methods can be expected to provide scalable and cost-effective options for offloading real-world testing. 

Nevertheless, employing neural rendering techniques for AV testing gives rise to a fundamental concern: How reliably can conclusions drawn from simulated data be transferred to real data? 
To address this concern, it is essential to assess whether the system, trained on real data, interprets real and simulated data similarly, as highlighted in \cref{fig:front-figure}. This divergence, termed the \emph{real2sim} gap \cite{yang2023unisim}, has received far less attention than its counterpart \emph{sim2real} gap, which pertains to transferring knowledge acquired in simulation to the real world \cite{hu2023simulation}. Traditionally, the real2sim problem has been addressed by improving the realism of rendered images. 
However, it is unknown how well common novel view synthesis (NVS) performance measures, such as PSNR, LPIPS~\cite{zhang2018unreasonable}, SSIM~\cite{wang2004image}, or FID~\cite{heusel2017gans}, correlate with a small real2sim gap, making it difficult to state what quality a given NeRF must reach to be useful for AV testing. 
Yet another aspect is that we are typically interested in the rendering quality when deviating from the original trajectory, \ie, in a setting where such metrics cannot be calculated due to the lack of ground truth data to compare with. 

In this paper, we propose a novel perspective on reducing the gap between real and simulated data for different perception modules of an autonomous system. Rather than improving upon the rendering quality, we aim to make the perception models more robust to NeRF artifacts without degrading performance on real data. We believe this direction to be complementary to increasing NeRF performance, and a potential key for making scalable, virtual AV testing a reality. As a first step in this direction, we show that even simple data augmentation techniques can have a large effect on model robustness against NeRF artifacts. 


Further, we perform the first extensive real2sim gap study on a large-scale AD dataset and assess the performance of three object detectors alongside an online mapping model on both real data and data from a state-of-the-art (SOTA) neural rendering method. 
Our investigation encompasses the impact of diverse data augmentation techniques during training, as well as the fidelity of NeRF renderings during inference. 
We find that integrating such data during model fine-tuning notably enhances their robustness to simulated data and, in some cases, even elevates performance on real data. 
Lastly, we investigate the correlation between the real2sim-gap and image reconstruction metrics to provide insights into what matters for applying NeRFs as simulators for AD data. We find LPIPS and FID to be strong indicators of the real2sim-gap, and that our proposed augmentations reduce the sensitivity to poor view synthesis. 
\section{Related work}
\label{sec:related_work}

\parsection{Novel view synthesis for autonomous driving}
NeRFs have emerged as a promising approach for simulating AD data. In contrast to game engine-based methods, NeRFs remove the need for manual asset creation and are optimized to create sensor-realistic renderings by design. 
However, a key challenge for NeRFs is handling the scale and dynamics of automotive scenes.
Neural Scene Graphs \cite{ost2021neural}, Panoptic Neural Fields \cite{kundu2022panoptic} and Panoptic NeRF \cite{fu2022panoptic} separate the background from moving actors by modeling each component with a separate, rigid, multi-layer perceptron (MLP). Still, these methods struggle with scaling to large scenes due to the limited expressiveness of the MLP. 
S-NeRF \cite{xie2023snerf} addresses this by building upon Mip-NeRF 360 \cite{barron2022mip360} to better handle unbounded scenes. However, its long training time makes it impractical to simulate many scenes. 
MARS \cite{wu2023mars} and UniSim~\cite{yang2023unisim} utilize the hash-grid representation from iNGP~\cite{muller2022instant} and achieve efficient models, although with limitations on sensor configuration. 
NeuRAD \cite{tonderski2023neurad} introduces efficient ways of modeling the important aspects of AD data, and achieves state-of-the-art performance across five AD datasets \cite{caesar2020nuscenes,Argoverse2,geiger2013vision,pandaset,alibeigi2023zenseact}. Some works in this domain~\cite{yang2023unisim,tonderski2023neurad,wu2023dynamic} make efforts to tailor their evaluation to closed-loop AD simulation. Nonetheless, their testing sets are small in the context of AD perception, and their applicability to downstream tasks at a larger scale is unexplored. Studies with large simulated test sets have thus far exclusively been done using game engine-based simulators, see for instance \cite{klinghoffer2023towards}, which compared to NeRFs require laboursome manual asset creation.

\parsection{Perception for autonomous driving}
Perception in 3D is a critical component of many autonomous driving solutions. Due to cameras' low cost and high availability, camera-only methods have been the subject of extensive research in recent years. 
For 3D object detection, FCOS3D~\cite{wang2021fcos3d} is a one-stage monocular object detector, building upon the 2D object detector FCOS~\cite{tian2020fcos}, but regressing targets in 3D rather than 2D.
PETR~\cite{liu2022petr} further supports the multi-view setting and instead builds upon the decoder architecture from DETR~\cite{carion2020end}, adding encoding points in the camera frustum into the image features. 
BEVFormer~\cite{li2022bevformer} also adopts a query-based architecture for multi-view input but encodes features into a bird's eye view (BEV) representation instead. 

In addition to detecting objects in 3D, many autonomous vehicles also estimate the road elements of their surroundings, also known as online mapping~\cite{li2022hdmapnet,liao2023maptr,liu2023vectormapnet,liao2023maptrv2,yuan2024streammapnet}. 
MapTRv2 \cite{liao2023maptrv2} is a current SOTA method that uses a vector-based representation for detected road elements such as lane dividers and road boundaries. 
Similarly to BEVFormer, MapTRv2 encodes image features and lifts them into a BEV representation. The objects' class and geometry are estimated through a DETR-like \cite{carion2020end} transformer decoder. 

\parsection{Domain adaptation and multi-task learning}
Domain adaptation (DA) is a field aiming to cope with issues arising when the training and evaluation data come from different distributions~\cite{farahani2021brief,liu2022deep}. 
Although different DA definitions exist, unsupervised DA is the most commonly studied version and assumes no access to labels in the target domain~\cite{wilson2020survey}. 
When labels exist in both the source and target domain, the problem can be categorized both as supervised domain adaptation~\cite{wilson2020survey} and multi-task learning~\cite{pan2009survey, zhang2021survey}. Multiple works study how to learn different tasks, such as semantic segmentation and depth estimation, simultaneously \cite{navon2022multi, khattar2021cross, Ishihara_2021_CVPR}. 
Our setting, however, differs from both these aspects, as the labels for the source and target domains are \textit{identical}. Before neural rendering techniques, these situations rarely arose, and therefore have only been scarcely studied~\cite{yang2023unisim}.
\begin{figure}[t]
    \centering
    \includegraphics[width=0.8\linewidth]{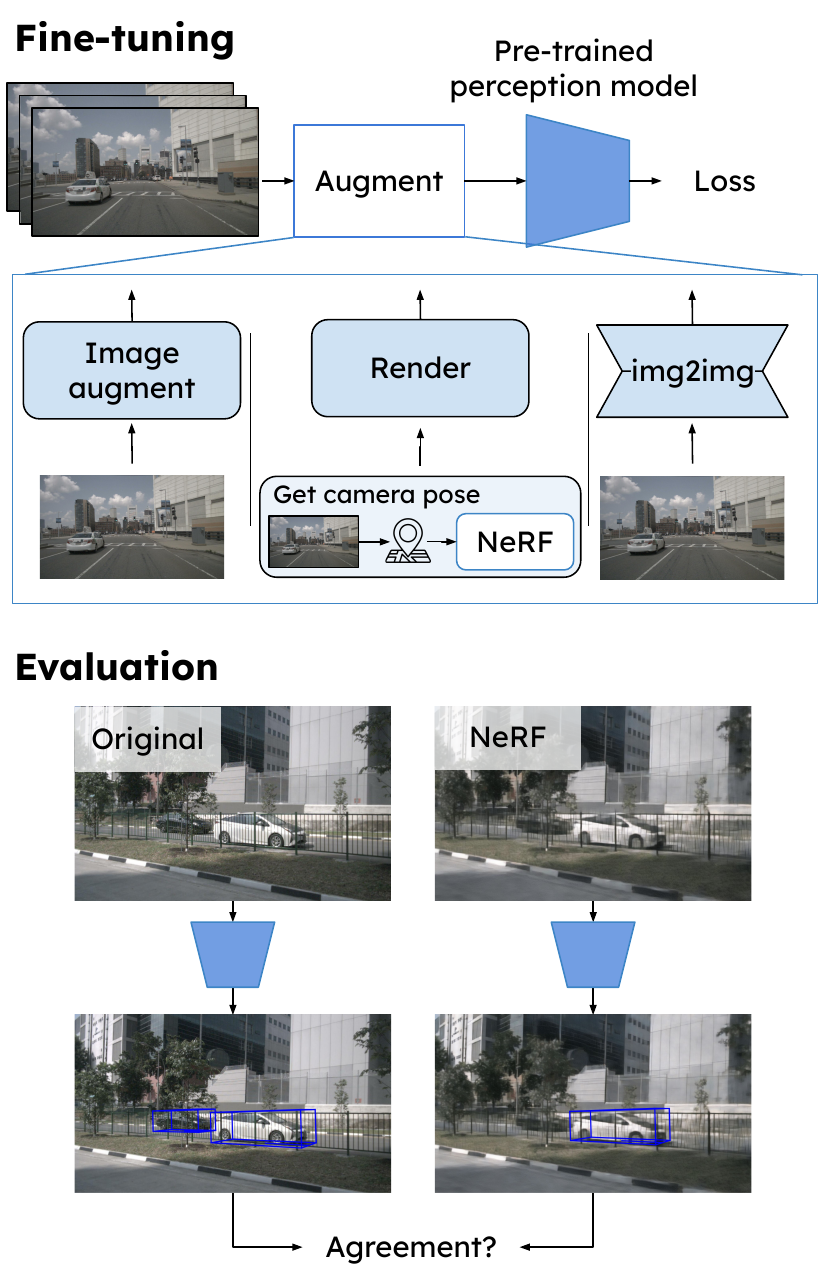}
    \caption{Overview of our data pipeline for fine-tuning (top) and evaluating (bottom) perception models. We explore three different augmentation methods for fine-tuning.}
    \label{fig:method}
    \vspace{-18pt}
\end{figure}
    
\begin{figure*}[t]
    \centering
         \centering
         \begin{subfigure}[b]{0.24\textwidth}
             \centering
             \includegraphics[width=\textwidth]{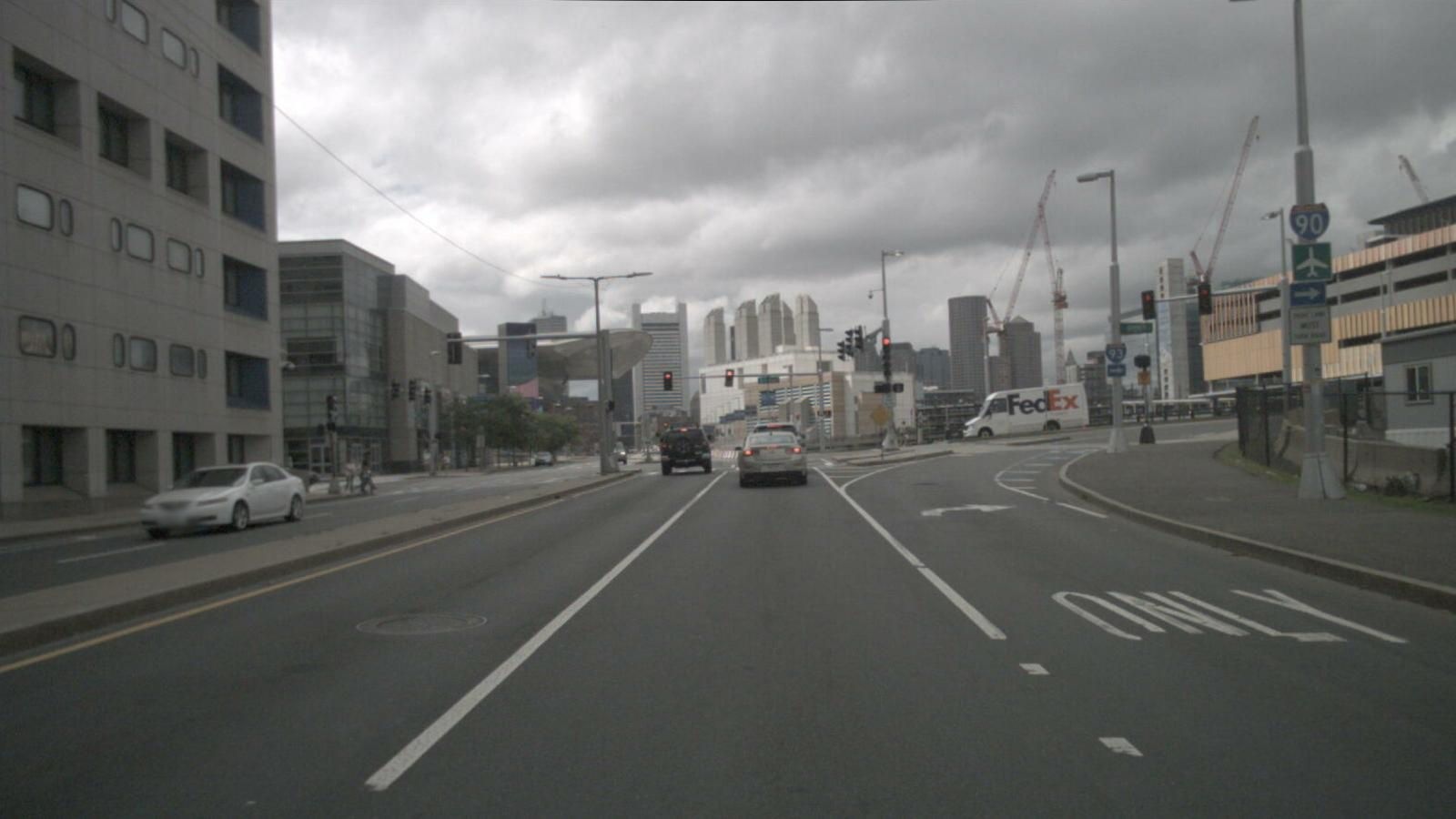}
             \caption{Original}
         \end{subfigure}
         \hfill
         \begin{subfigure}[b]{0.24\textwidth}
             \centering
             \includegraphics[width=\textwidth]{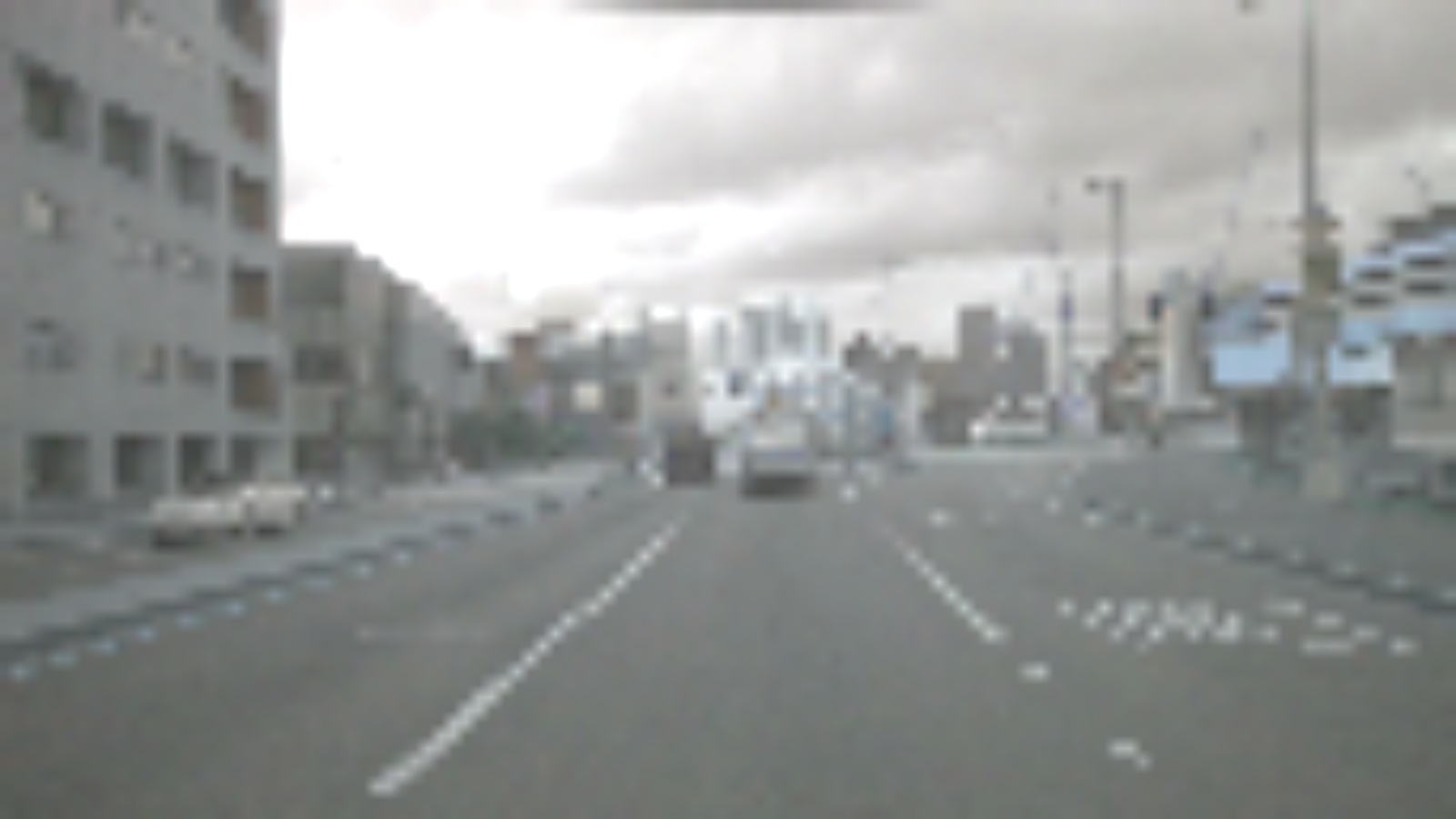}
             \caption{Image Augmentation}
         \end{subfigure}
         \hfill
         \begin{subfigure}[b]{0.24\textwidth}
             \centering
             \includegraphics[width=\textwidth]{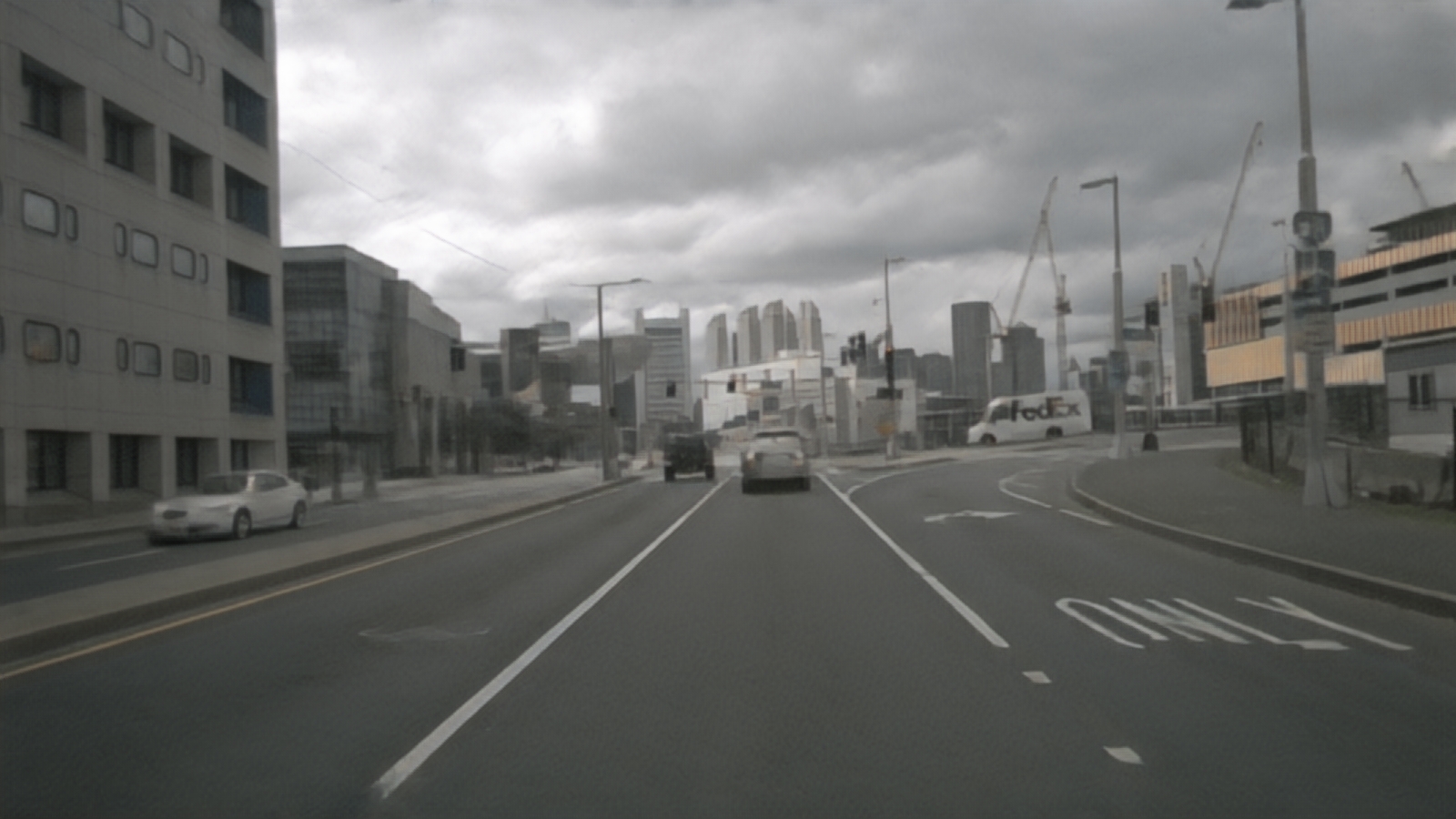}
             \caption{NeRF}
         \end{subfigure}
         \hfill
         \begin{subfigure}[b]{0.24\textwidth}
             \centering
             \includegraphics[width=\textwidth]{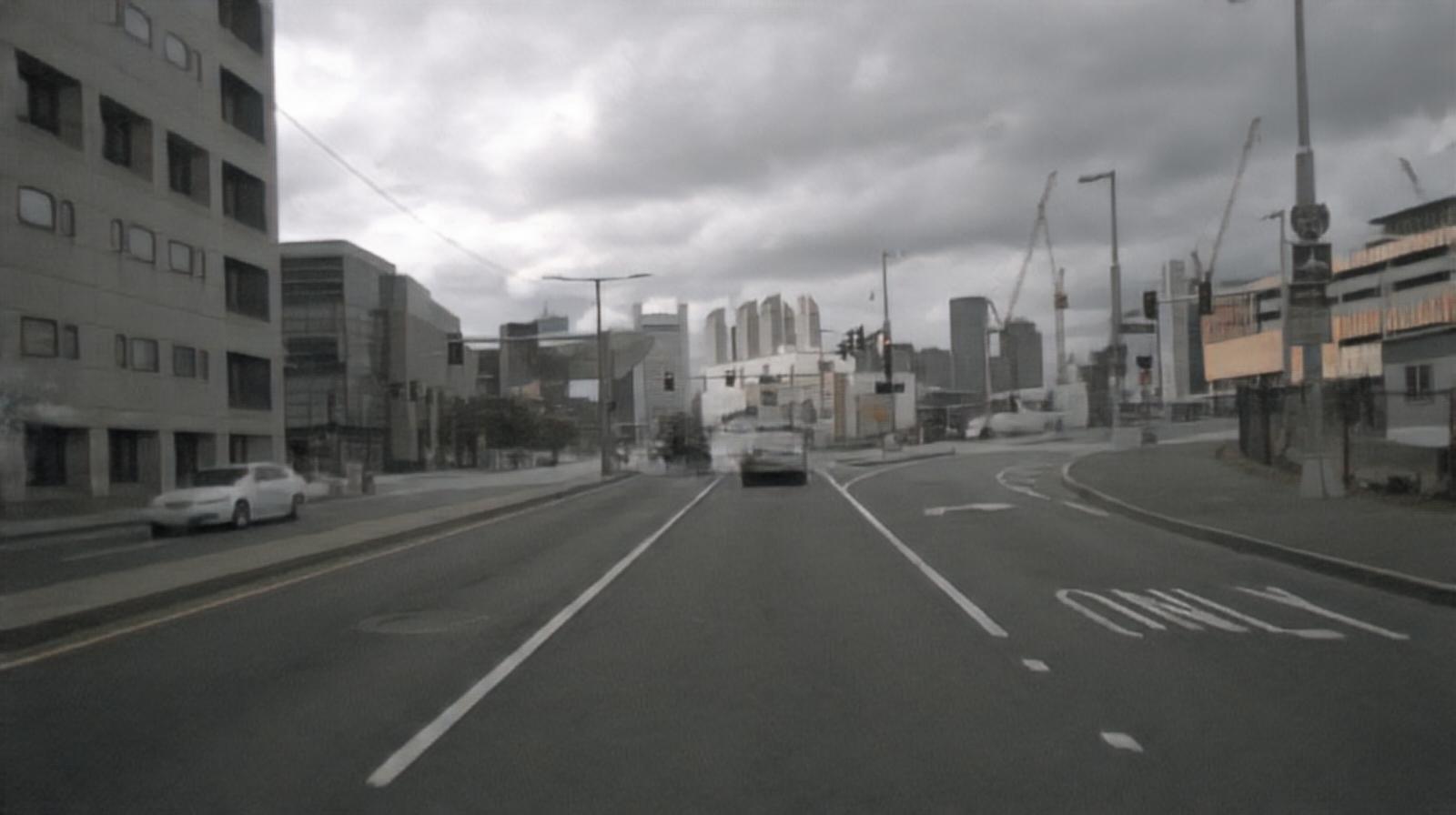}
             \caption{Img2Img}
         \end{subfigure}

         \begin{subfigure}[b]{0.24\textwidth}
             \centering
             \includegraphics[width=\textwidth]{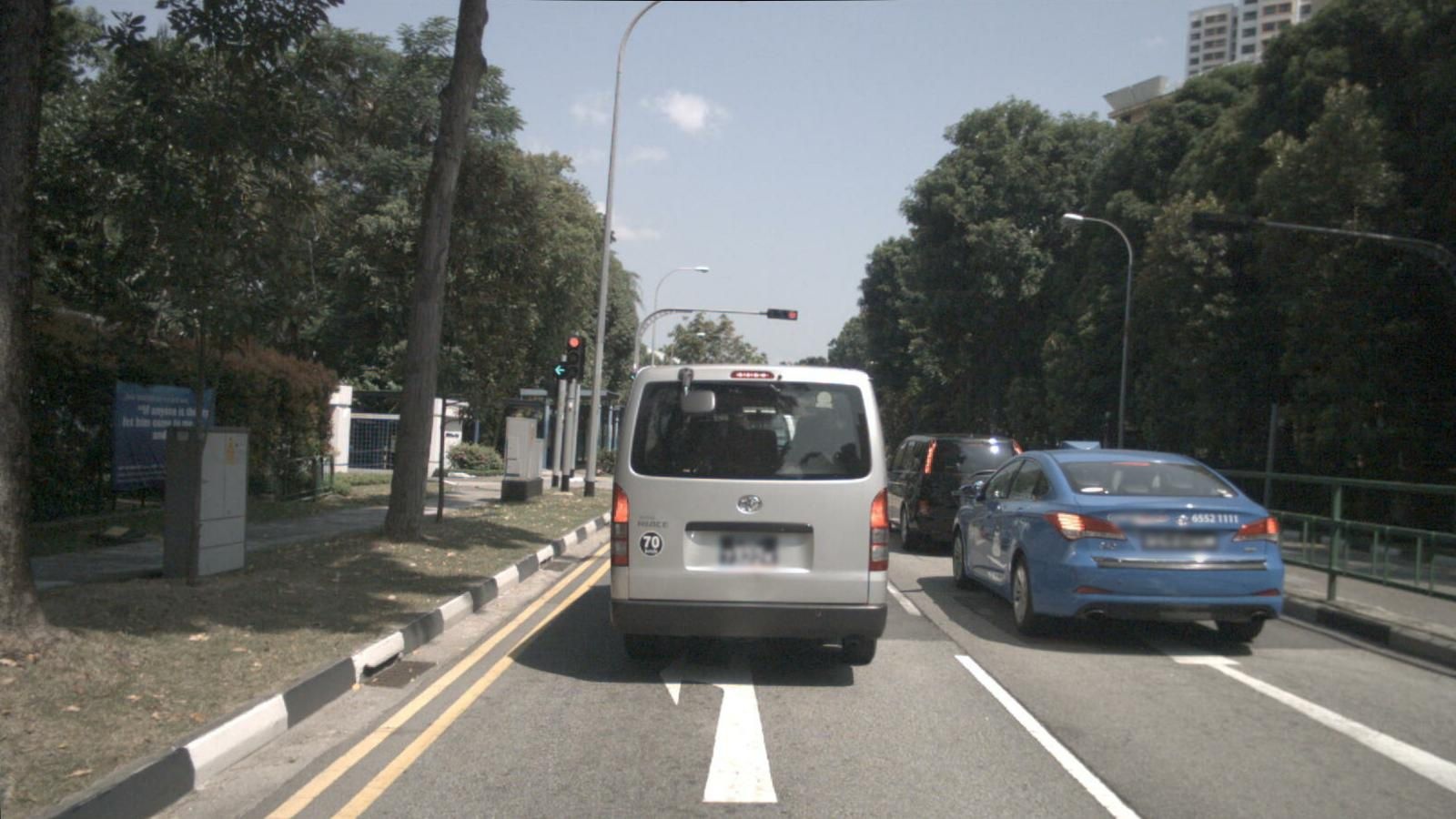}
         \end{subfigure}
         \hfill
         \begin{subfigure}[b]{0.24\textwidth}
             \centering
             \includegraphics[width=\textwidth]{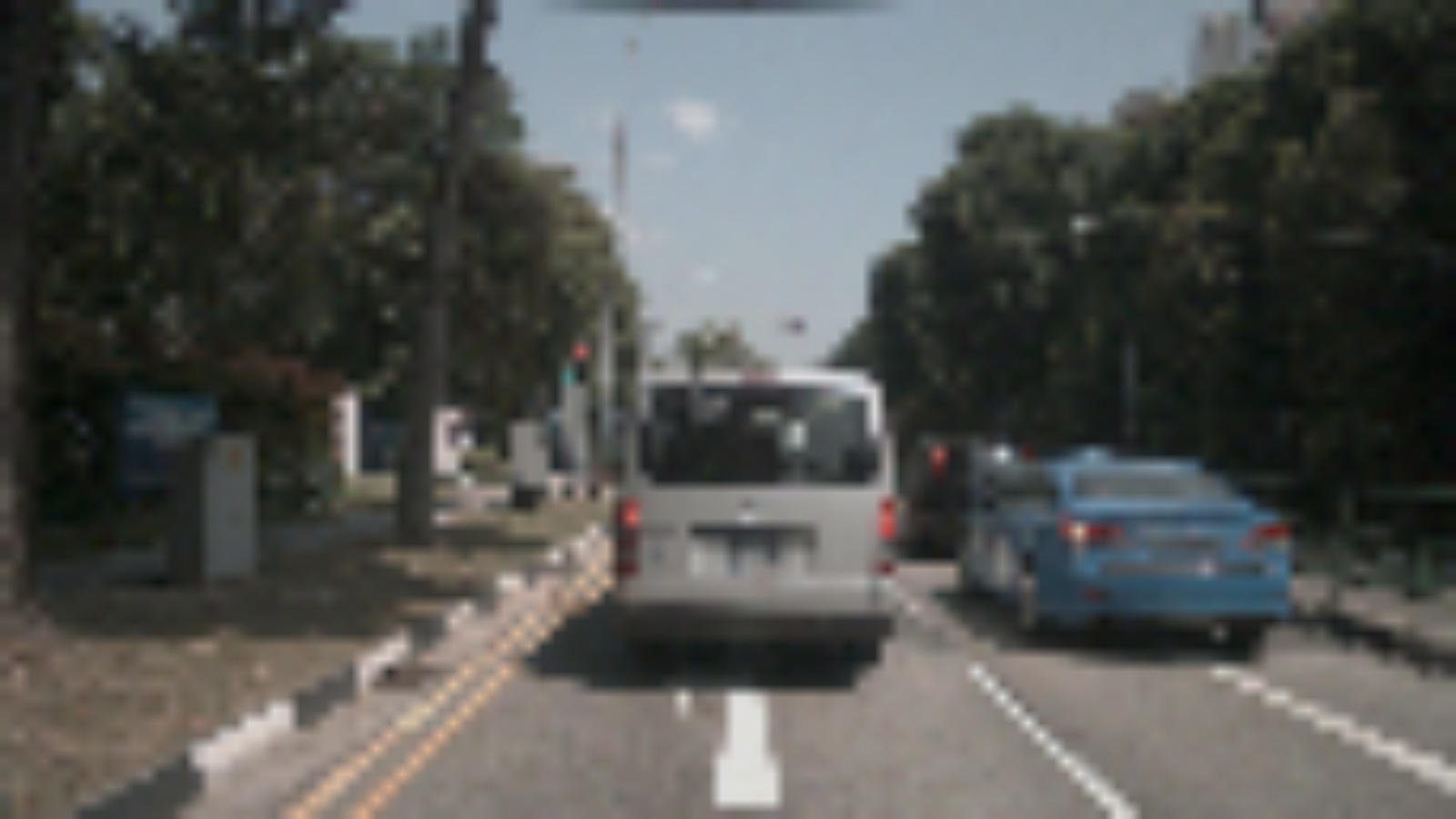}
         \end{subfigure}
         \hfill
         \begin{subfigure}[b]{0.24\textwidth}
             \centering
             \includegraphics[width=\textwidth]{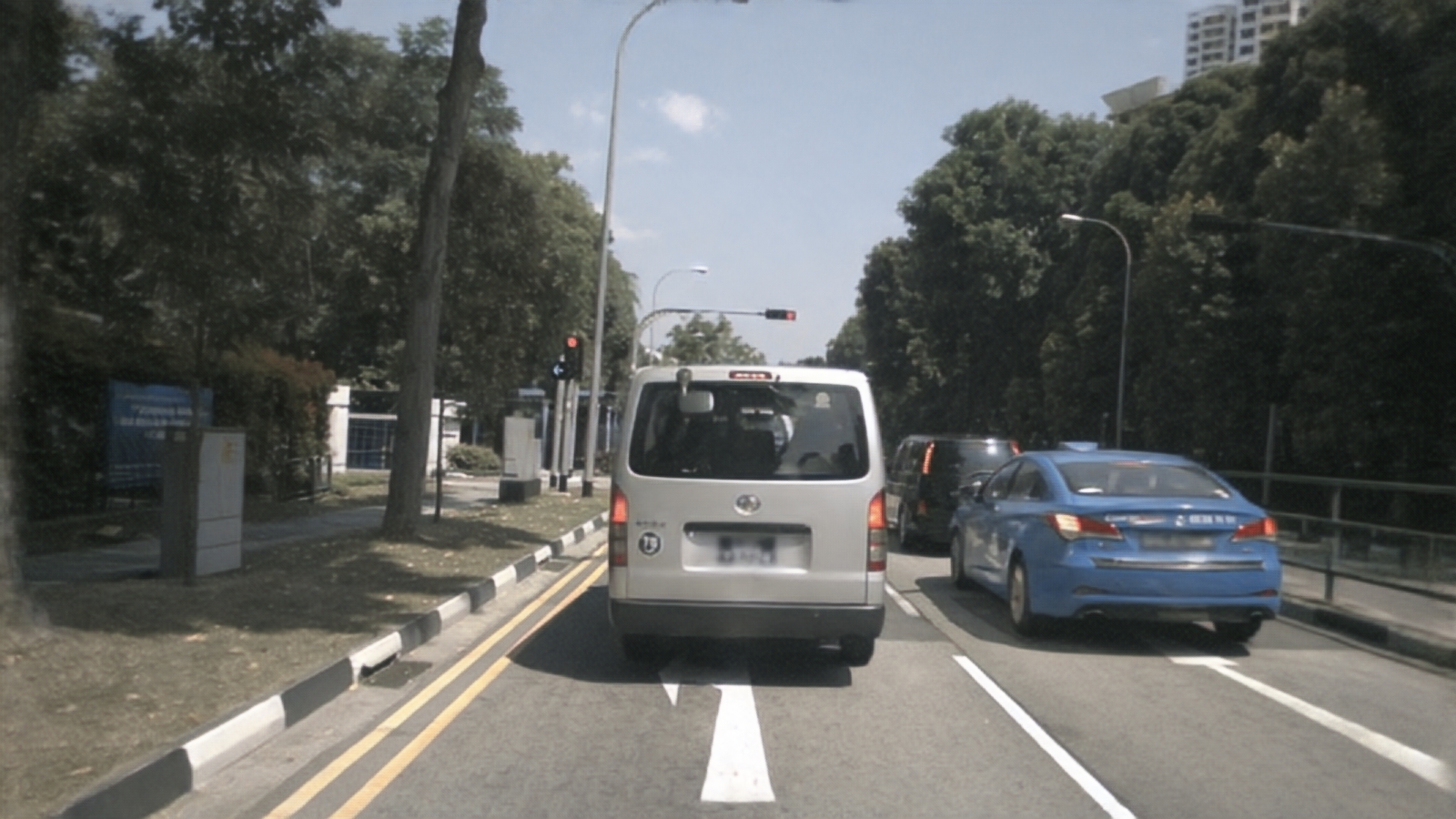}
         \end{subfigure}
         \hfill
         \begin{subfigure}[b]{0.24\textwidth}
             \centering
             \includegraphics[width=\textwidth]{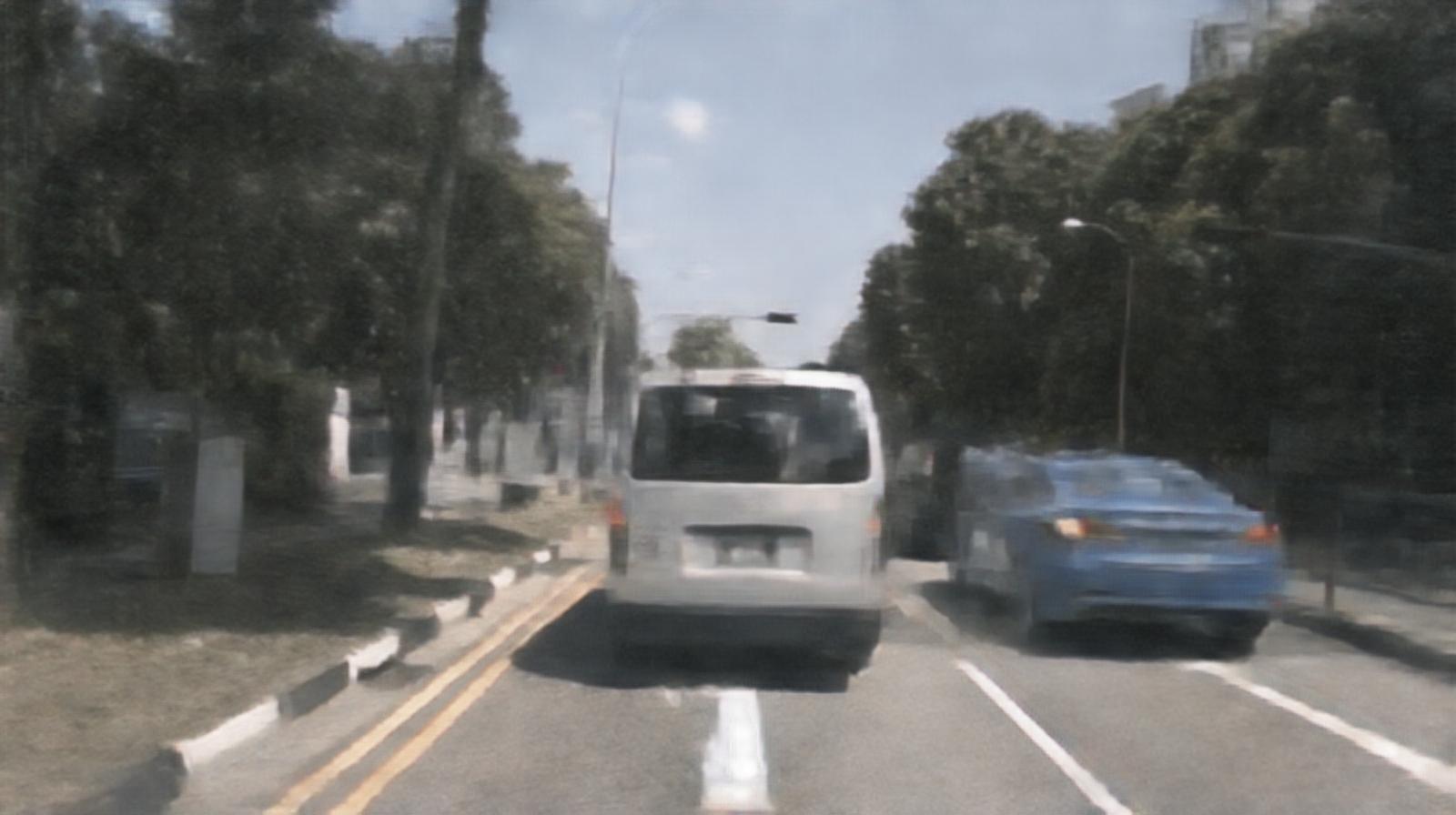}
         \end{subfigure}
    \caption{Examples of our different data augmentation strategies to make perception models more robust.}
    \label{fig:data_aug_examples}
    \vspace{-18pt}
\end{figure*}

\section{Method}
\label{sec:method}


Our goal is for AD systems to behave the same way when exposed to rendered and real data. As a first step in this direction, we explore how different fine-tuning strategies can make perception models more robust to artifacts in the rendered data. Specifically, given already trained models, we fine-tune the perception models using images designed to improve performance on rendered images while maintaining performance on real data, see \cref{fig:method}. Besides reducing the real2sim gap, this can potentially lower requirements on sensor-realism, opening for a wider applicability of neural rendering methods, and lessening computational needs for training and evaluating of the said methods. Note that, while we focus on perception models, our methodology can easily be extended to end-to-end models as well~\cite{hu2023planning,jiang2023vad}.

Last, we acknowledge that one can imagine multiple ways to achieve the goal of making models more robust, for instance by drawing inspiration from domain adaptation \cite{liu2022deep,farahani2021brief} and multi-task learning \cite{zhang2021survey} literature. However, fine-tuning requires minimal model-specific adjustments, allowing us to study a range of models easily.

\subsection{Image augmentations}
\label{sec:method_img_aug}
A classic strategy to obtain increased robustness to artifacts is to use image augmentations~\cite{simard2003best,murphy2022probabilistic}. Here, we select augmentations to represent various distortions present in rendered images. More specifically, we add random Gaussian noise, convolve the image with a Gaussian blur kernel, apply photometric distortions similar to the ones found in SimCLR~\cite{chen2020simple}, and, finally, downsample and upsample the image. The augmentations are applied sequentially, each with some probability. For reference, the perception models considered in this work are generally trained with no augmentations affecting image quality, or only photometric distortions. Details on hyperparameters can be found in the supplementary material, \cref{sec:img_aug_hyperparameters}.

\subsection{Fine-tuning with mixed-in rendered images}
\label{sec:method_nerf_aug}
Another natural way to adapt perception models to NeRF-rendered data is to include such data during fine-tuning. This involves training a NeRF method on the same dataset used to supervise perception models $\mathcal{D}_\text{train}^\text{real}$. However, training NeRFs on all of $\mathcal{D}_\text{train}^\text{real}$ can be prohibitively expensive for large datasets. Instead, we train NeRFs on a subset $\mathcal{D}^\text{nerf} \subset \mathcal{D}_\text{train}^\text{real}$. 
Note that in addition to annotations for the given perception task, NeRFs for AD typically add the requirement of data in $\mathcal{D}^\text{nerf}$ to be sequential, where some additionally require labels for tasks such as 3D object detection~\cite{yang2023unisim,tonderski2023neurad}, semantic segmentation~\cite{kundu2022panoptic}, or multiple types of labels~\cite{wu2023mars}.

Next, we divide the images for the selected sequences in $\mathcal{D}^\text{nerf}$ into NeRF training $\mathcal{D}^\text{nerf}_\text{train}$ and holdout $\mathcal{D}^\text{nerf}_\text{fine-tune}$ sets. Fine-tuning of the perception models is done on their entire training dataset $\mathcal{D}_\text{train}^\text{real}$, and for images that have a rendered correspondence in $\mathcal{D}^\text{nerf}_\text{fine-tune}$, we use the rendered images with probability $p$. This implies that the images utilized for fine-tuning have not been seen by the NeRF model.

\subsection{Image-to-image translation}
\label{sec:method_pix2pix}
As mentioned previously, rendering NeRF data is an expensive data augmentation technique. Furthermore, it requires sequential data and potentially additional labeling beyond what is needed for the perception task. That is, to obtain a scalable method, we would ideally like an efficient strategy to obtain NeRF data for single images. To this end, we propose to learn to generate NeRF-like images using an image-to-image method. Given a real image, the model translates the image to the NeRF domain, effectively introducing typical NeRF artifacts. This enables us to vastly increase the amount of NeRF-like images during fine-tuning at a limited computational cost. We train the image-to-image model~\cite{wang2018high} using rendered images $\mathcal{D}^\text{nerf}_\text{fine-tune}$ and their corresponding real images. See \cref{fig:data_aug_examples} for visual examples of the different augmentation strategies.

\section{Results}
\label{sec:results} 
In this section, we first describe our experimental setting. We then show how the real2sim gap differs for different perception models, tasks, and augmentation strategies. Next, we study how the models with the smallest gap perceive renderings with large changes in viewpoint where no corresponding real images have been collected. Finally, we show the correlation between common metrics for novel view synthesis and the real2sim gap. 

\subsection{Experimental setup}
We assess the real2sim gap by comparing the performance of multiple perception models when applied to real data versus their NeRF-rendered counterparts. We validate our approach to improving the robustness of perception models by implementing various augmentation strategies during fine-tuning of the models, as detailed in \cref{sec:method}. Striving for a general approach adaptable to diverse models and tasks, we employ the same augmentation techniques across all models and tasks. Given the critical need to not sacrifice real-world performance, we also assess our augmentation methods on real data. Specifically, we test the three augmentation techniques described in \cref{sec:method_img_aug,sec:method_nerf_aug,sec:method_pix2pix}; traditional image augmentations, rendered data, and image-to-image translation. See \cref{sec:nerf_aug_details} respectively \cref{sec:image_to_image_aug_details} for details regarding the datasets used during fine-tuning, and \cref{sec:image_to_image_train_details} for details on training the image-to-image model. 

\parsection{Neural rendering}
We opt to employ NeuRAD~\cite{tonderski2023neurad}, the current SOTA method for NVS on AD data, as our NeRF-method. NeuRAD relies on a hashgrid representation~\cite{muller2022instant} and upsampling techniques for efficient learning and inference of high-resolution data. To handle dynamic scenes, it models all actors to be rigid entities, which limits expressiveness for deformable actors such as pedestrians. Nonetheless, NeuRAD is currently the most performant NVS technique across multiple AD datasets. For the perception validation sequences, we train NeuRAD on all images except those held out for evaluating the perception models. For reference, statistics for the NeuRAD trainings are shown in \cref{tab:neurad_results}, where we observe similar performance on images used for data augmentation as the ones used for evaluating the real2sim gap.

\parsection{Dataset and tasks}
Our evaluations are conducted on nuScenes, a widely recognized research ground for perception models. Compared to other large AD datasets such as Waymo Open Dataset~\cite{sun2020scalability}, Argoverse2~\cite{Argoverse2} or Zenseact Open Dataset~\cite{alibeigi2023zenseact}, nuScenes uses lower resolution cameras and lidars, posing an interesting challenge for neural rendering methods. We evaluate all perception models on a subset of the official validation split, namely scenes collected at daytime and without heavy rain. This is because lens flares and water spray cannot currently be handled by any neural rendering method for AD data. This filtering results in 111 scenes used for evaluation, see \cref{sec:eval_scenes} for details.

Further, we evaluate two tasks: 3D object detection and online mapping. These tasks consider complementary aspects of the real2sim gap, as the former focuses on foreground objects, while the latter targets the static background. Both tasks and corresponding models are described in more detail below.

\parsection{3D object detection}
We evaluate the gap across three camera-only 3D object detection (3DOD) models, chosen to represent different aspects of prevalent model architectures. Namely, we apply FCOS3D~\cite{wang2021fcos3d}, a fully-convolutional monocular detector, PETR~\cite{liu2022petr}, a multi-view and 3D adaptation of DETR~\cite{carion2020end}, and BEVFormer~\cite{li2022bevformer} a multi-view detector centered around the bird's-eye-view representation. We follow the evaluation protocol established by the nuScenes object detection task and report mean Average Precision (mAP) and nuScenes Detection Score (NDS) on both real and rendered data. Additionally, we analyze the consistency between detections made on real and rendered data. This way, we do not only consider performance in absolute terms, but also measure if the model makes the same mistakes on both types of data. To this end, we compute NDS twice, with a distance threshold of $2$\si{m}, each time treating the other set of detections as ground truth. We average the results from both evaluations to get our detection agreement (DA). All models are initialized from weights pre-trained on nuScenes and fine-tuned for a fixed number of gradient steps. See \cref{sec:3dod_training_details} for more details on model weights and hyperparameters used for fine-tuning.  

\parsection{Online mapping}
To extend our evaluations beyond the 3D object detection task, we also evaluate MapTRv2~\cite{liao2023maptrv2} on the task of online mapping. Following the evaluation framework outlined in \cite{lilja2023localization}, we compute mAP for the classes “divider”, “boundary”, and “crossing”. We also compute detection agreement in the same fashion as for 3DOD, alternating which set of detections is used as ground truth. 

For completeness, we use both the original validation split (same as for 3DOD) and the geographically disjoint split recently proposed in~\cite{lilja2023localization}. In short, the original split suffers from data leakage, as there is significant geographical overlap between training and validation/testing samples. As an effect, generalization performance is largely overestimated when using the original split. Note that for the geographically disjoint split, we again remove scenes at night or with rain, resulting in 154 scenes used for evaluation. See \cref{sec:eval_scenes} for details. 

\subsection{Real2sim gap on interpolated views}
We begin with studying the gap on interpolated views. These viewpoints lie in between images used to supervised NeuRAD and have corresponding real images. While these views arguably are easier to render than, for instance, shifts in the ego-vehicle position, they allow a direct comparison between real and simulated data. 
The real2sim gaps for 3DOD and online mapping models are reported in \cref{tab:real2sim-results-3dod}, and discussed in detail below. For all metrics, the gap is expressed as the relative performance drop compared to the real-world performance of the model without augmentations. 
\begin{table*}[t]
    \small
    \centering
    \caption{Real2sim results for the 3D object detection models and the online mapping method MapTRv2, fine-tuned with different strategies. ''Sim`` indicates that the model was evaluated on rendered data from NeuRAD. For online mapping, Original and Geographically Disjoint (Geogr.) refers to the splits used for training and evaluation. }
    \setlength{\tabcolsep}{3pt} 
    \setlength\extrarowheight{-5pt}
    \begin{tabular}{c c ccc | ccc | crc | cc | rc}
    \hline 
    \toprule 
    \multirow{3}{*}{\begin{tabular}{c}Fine-tuning \\ method\end{tabular}} & \multirow{3}{*}{\begin{tabular}{c}Evaluation \\ data\end{tabular}} & \multicolumn{9}{c}{3D object detection} & \multicolumn{4}{c}{Online mapping} \\ \cmidrule(lr){3-11} \cmidrule(lr){12-15}
    & & \multicolumn{3}{c}{FCOS3D}                           & \multicolumn{3}{c}{PETR}                            & \multicolumn{3}{c}{BEVFormer}  & \multicolumn{2}{c}{Original} & \multicolumn{2}{c}{Geogr.}             \\  
                                  &                       & mAP             & NDS             & DA               & mAP             & NDS             & DA              & mAP             & NDS             & DA                 & mAP    & DA     & mAP    & DA  \\ \midrule
    \multirow{3}{*}{Real data only}
                                  & Real                 & $32.2$          & $39.8$          &                  & $38.6$          & $43.1$          &                 & $38.4$          & $48.5$           &                    & $64.5$ &        & $26.6$ &    \\ 
                                  & Sim                  & $13.5$          & $28.8$          & $46.3$           & $20.2$          & $31.6$          & $55.4$          & $29.1$          & $42.7$           & $76.6$             & $54.0$ & $71.8$ & $23.2$ & $67.1$ \\ 
                                  & Gap (\%) $\downarrow$             & $58.1$          & $27.6$          & $53.7$           & $47.7$          & $26.7$          & $44.6$          & $24.2$          & $12.0$           & $23.4$             & $16.3$ & $28.2$ & $12.8$ & $32.9$ \\ \midrule
    \multirow{3}{*}{\begin{tabular}{c}Image \\ augmentations\end{tabular}}  
                                  & Real                  & $32.5$          & $40.0$          &                  & $34.0$          & $38.9$          &                 & $38.9$          & $48.6$          &                    & $65.1$ &        & $26.5$ &     \\
                                  & Sim                   & $13.5$          & $28.9$          & $46.5$           & $20.4$          & $30.0$          & $57.6$          & $31.0$          & $44.0$          & $77.6$             & $55.2$ & $72.3$ & $23.8$ & $70.1$ \\ 
                                  & Gap (\%) $\downarrow$             & $58.1$          & $27.4$          & $53.5$           & $47.2$          & $30.4$          & $42.4$          & $19.3$          & $9.3$           & $22.4$             & $14.4$ & $27.7$ & $10.5$ & $29.9$ \\ \midrule
    \multirow{3}{*}{NeRF}         & Real                  & $31.2$          & $38.6$          &                  & $35.1$          & $40.0$          &                 & $38.5$          & $48.3$          &                    & $64.9$ &        & $26.9$ &    \\
                                  & Sim                   & $23.5$          & $33.6$          & $58.7$           & $29.3$          & $37.3$          & $70.7$          & $31.7$          & $44.5$          & $78.9$             & $56.0$ & $74.5$ & $24.6$ & $70.5$ \\ 
                                  & Gap (\%) $\downarrow$             & $27.0$          & $15.6$          & $\mathbf{41.3}$  & $\mathbf{24.1}$ & $\mathbf{13.5}$ & $\mathbf{29.3}$ & $17.4$          & $8.2$           & $21.1$             & $13.2$ & $25.5$ & $\mathbf{ 7.5}$ & $29.5$\\ \midrule
    \multirow{3}{*}{Image-to-image}      
                                  & Real                  & $32.5$          & $39.8$          &                  & $31.4$          & $37.2$          &                 & $37.5$          & $48.1$          &                    & $62.5$ &        & $25.2$ &    \\ 
                                  & Sim                   & $24.5$          & $34.3$          & $57.3$           & $26.1$          & $35.1$          & $67.9$          & $33.0$          & $44.9$          & $80.7$             & $56.8$ & $74.8$ & $24.3$ & $73.9$\\ 
                                  & Gap (\%) $\downarrow$             & $\mathbf{23.9}$ & $\mathbf{13.8}$ & $42.7$           & $32.4$          & $18.6$          & $32.1$          & $\mathbf{14.1}$ & $\mathbf{7.4}$  & $\mathbf{19.3}$    & $\mathbf{11.9}$ & $\mathbf{25.2}$ & $ 8.3$ & $\mathbf{26.1}$\\

    \bottomrule 
    \end{tabular}
    \label{tab:real2sim-results-3dod}
    \vspace{-10pt}
\end{table*}

\parsection{Gap for models without fine-tuning}
Despite leveraging the current SOTA in NVS for AD data, \cref{tab:real2sim-results-3dod} shows a significant real2sim gap across all models and tasks. Notably, there is a considerable variation in the gap among different 3DOD models, with BEVFormer exhibiting the smallest gap, whereas the mAP performance of FCOS3D is more than halved. Further, we observe a greater gap for mAP than for NDS. NDS is a weighted score where half of it consists of mAP, while the other half measures errors in terms of translation, scale, orientation, velocity, and attribute for true positives only. Considering this, the gap mainly stems from spurious or missing detection, while the quality of the true positives in terms of scale, orientation, etc., is less affected. For the online mapping task, we can see a smaller gap than for the 3DOD models, which is natural since NeuRAD often renders the static parts of a scene more accurately than the dynamic parts~\cite{tonderski2023neurad}. 


\parsection{Image augmentations}
The efficacy of basic image augmentations varies among the different models. Both BEVFormer and MapTRv2 demonstrate enhancements on simulated data while maintaining or improving performance on real data. However, FCOS3D exhibits minimal to no improvement on simulated data, despite enhancing performance on real data. In contrast, PETR displays a larger gap in terms of NDS, along with a significant decline in performance on real data. Analyzing the detection agreement shows improved consistency across the board, albeit with relatively modest improvements for FCOS3D.

\parsection{Rendered data}
Incorporating rendered images during fine-tuning decreases the gap across all models. FCOS3D and PETR demonstrate substantial improvements of $74.1$\% respectively $45.5$\% in mAP on simulated data, with slight degradations on real data. Additionally, the models fine-tuned on rendered data show significant improvements in terms of detection agreement, indicating a higher consistency to the real-world detections.

\parsection{Image-to-image translation}
Finally, fine-tuning with image-to-image translated images leads to a significant increase in performance on simulated data across all models. It is notable that this artificial extension with NeRF-like data performs better than using the actual NeRF-data for multiple methods. However, for most methods, this also comes with some penalty in mAP performance on real data. 

\begin{figure*}[htbp]
    \begin{subfigure}[b]{0.3\textwidth}
        \centering
        \includegraphics[width=\textwidth]{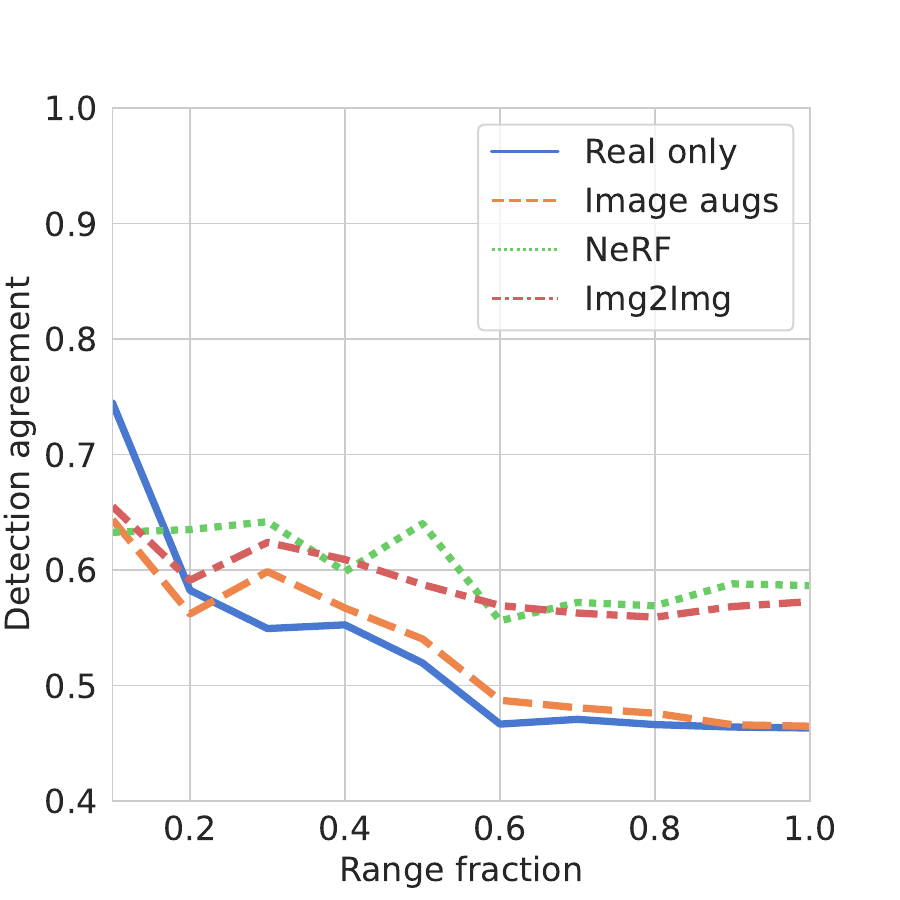}
        \caption{FCOS3D.}
        \label{fig:da-vs-range-fcos3d}
    \end{subfigure}
    \hfill
    \begin{subfigure}[b]{0.3\textwidth}
        \centering
        \includegraphics[width=\textwidth]{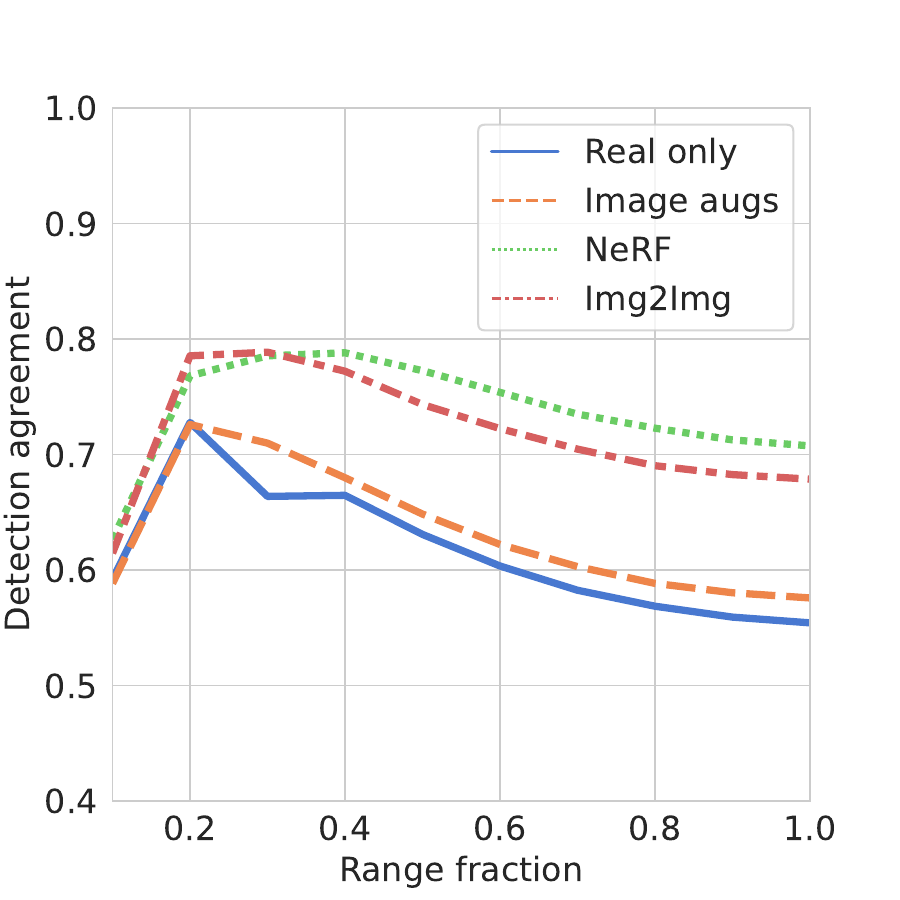}
        \caption{PETR.}
        \label{fig:da-vs-range-petr}
    \end{subfigure}
    \hfill
    \begin{subfigure}[b]{0.3\textwidth}
        \centering
        \includegraphics[width=\textwidth]{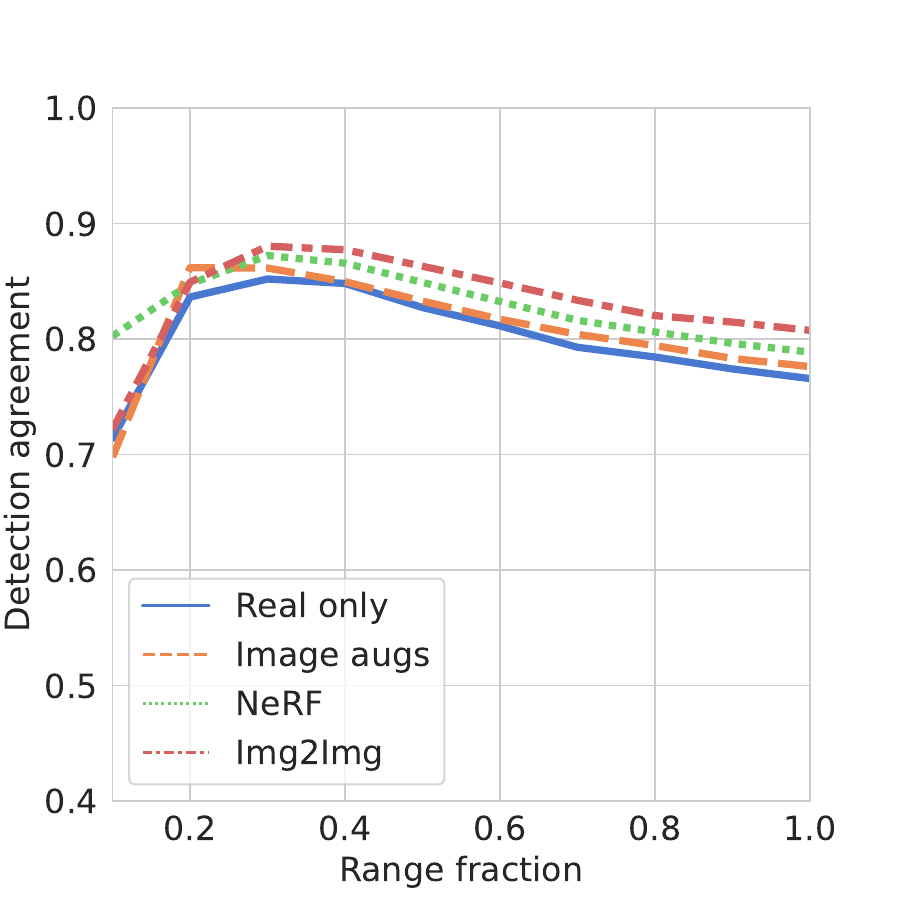}
        \caption{BEVFormer.}
        \label{fig:da-vs-range-bevformer}
    \end{subfigure}
    \caption{Detection agreement vs. fraction of the evaluation range, evaluated for the 3DOD models with different fine-tuning methods.}
    \label{fig:da-vs-range-3dod}
\end{figure*}

\parsection{Detection agreement across different distances}
To further gain insights into the consistency of detections relative to the detection distance, we assess the detection agreement across various fractions of the evaluation range used in the nuScenes protocol. Specifically, we examine the detection agreement for our 3D object detection models, with our different augmentations, across evaluation ranges ranging from $10$\% to $100$\% of the official protocol's evaluation range, which is 30--50m depending on the class. The results, illustrated in \cref{fig:da-vs-range-3dod}, reveal a decrease in detection agreement as evaluation distances increase, which aligns with the anticipated degradation in detection performance over longer distances. Interestingly, the fine-tunings with NeRF and image-to-image translated data notably reduce this effect for FCOS3D and PETR, as evidenced by an increasing disparity relative to the other augmentations. 

\subsection{Real2sim gap on extrapolated views}
\label{sec:real2sim_extrap}
To apply a NeRF in, for instance, closed-loop simulation or data generation, it needs to produce meaningful images not only for interpolated views but even more so when extrapolating to views away from the original trajectory. 
To address this, we render images when laterally shifting or rotating the ego-vehicle, effectively simulating scenarios such as the vehicle performing a lane change or having departed from its original lane. 
We evaluate the detection agreement and perception performance by adjusting the 3D labels and detections to accommodate the specified shift or rotation.
For clarity, we only evaluate models pre-trained with the image-to-image augmentation, as these displayed the best robustness in \cref{tab:real2sim-results-3dod}.

Compared to the interpolation setting, the extrapolation comes with unique challenges for evaluating the real2sim gap. First, the absence of real images in these novel scenarios prevents a comprehensive analysis of the true disparity. 
Instead, the field has traditionally relied on FID~\cite{NIPS2017_8a1d6947} as a performance measure on extrapolated camera views. Second, there are no assurances that the perception model would produce identical detections from the altered viewpoint, even with real data. By shifting the ego-vehicle, a scenario can become more challenging, \eg, by introducing partial occlusions, or the scenario can simply be less common in the collected data, \eg, images during a lane-shift. Thus, it is hard to completely disentangle these effects from the rendering quality. 

\begin{figure}[t]
    \centering
    \includegraphics[width=\linewidth]{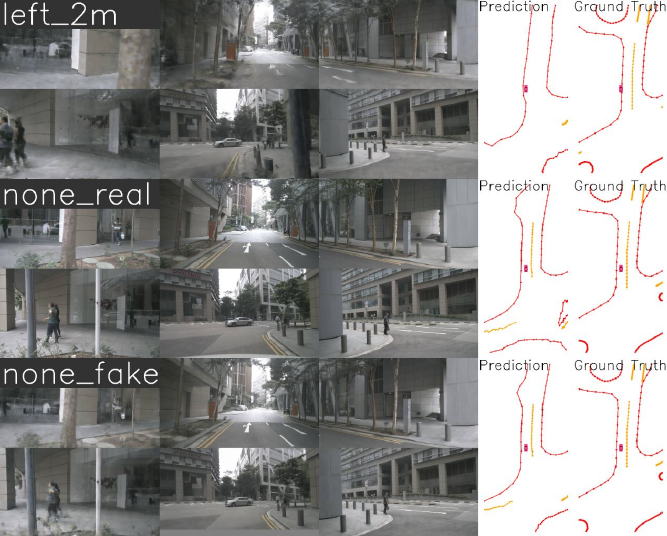}
    \caption{Online mapping predictions and ground truth for images shifted to the left (top), real images (middle), and rendered images without shift (bottom). When input data is shifted 2m to the left, the left road boundary, highlighted in red, should be straddled by the ego vehicle. However, the predictions maintain the ego vehicle within the boundary despite the image shift. }
    \label{fig:om-lane-shift-example}
\end{figure}

Following previous work~\cite{tonderski2023neurad,yang2023unisim} we render views when the ego vehicle has been moved $\pm 1$ and $\pm 2$ meters laterally. 
To ensure that the shifted views remain reasonable, \eg not inside other road users or structures, we select a smaller subset of scenes and manually validate the shifts' feasibility. See \cref{sec:eval_scenes} for further details. 
The FID score and perception performance on lateral shifts can be seen in \cref{tab:laneshifted-results-3dod} and \cref{tab:laneshifted-results-OM} for 3DOD and online mapping, respectively. 
While the performance of all object detection methods drops as the shift increases, the ranking among the methods persists. 
BEVFormer is the most robust, with NDS-score dropping only $5$ points from shifting the input data $2$m from the original position. 
For the online mapping method MapTRv2, the drops are relatively large. 
Even shifting the ego position $1$m yields a $13$\% and $18$\% drop on original and geographically disjoint splits, respectively. 
This discrepancy is surprising since the input data is the same as for BEVFormer and the architectures are fairly similar. 
By inspecting evaluation samples, we see that the model struggles with predicting rare training events, \eg, the vehicle traveling slightly outside the road as \cref{fig:om-lane-shift-example} exemplifies. 

For the rotation, the cameras are rotated around the ego-vehicle reference frame at discrete angles $\pm 5$\textdegree, $\pm 15$\textdegree, $\pm 30$\textdegree, $\pm 90$\textdegree ~and $180$\textdegree. As the resulting camera positions are expected to remain within, or close to, the ego vehicle's original extent, we here use the same validation sets as in \cref{sec:results}. Upon inspecting the renderings, they do not deteriorate noticeably with increasing rotation angle. In \cref{tab:maptrv2-rotation-finetuning} we see similar behavior as for the lateral shift. 
The online mapping performance under these rotations deteriorates for each angle as we rotate further from the original pose for both the original and geographically disjoint splits. 

\begin{figure}[t]
    \centering
    \includegraphics[width=\linewidth]{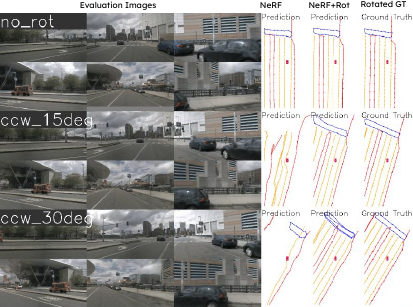}
    \caption{Online mapping predictions and ground truth for rotated input images, ccw denotes counter-clockwise. The predictions are greatly improved by injecting rotated scenarios during fine-tuning of the model.}
    \label{fig:om-rotation-example}
    \vspace{-15pt}
\end{figure}

Although the image quality degrades for larger rotations as indicated by the FID scores, we find the mapping performance to be overly sensitive to these viewpoint changes, as shown in \cref{fig:om-rotation-example}. We theorize that the drop in performance also stems from these scenarios being very rare in the training data. For instance, lane changes with harsh attack angles towards a lane marker are scarce compared to in-lane driving.

To this end, we fine-tune MapTRv2 with simulated rotated views of all angles on the training data inserted into the full training set.
As \cref{tab:maptrv2-rotation-finetuning} depicts, the performance on rotated views can be improved substantially by incorporating simulated such scenarios also during training. 
For instance, the performance using the geographically disjoint split improves from $5.2$ to $17.3$ mAP on novel views perpendicular to the original poses. This is also reflected in the predictions visualized in \cref{fig:om-rotation-example}.
Further, it is notable that the performance on real data is improved from $26.6$ to $27.5$ mAP. 
This indicates that utilizing NeRF-rendered data also for training could be beneficial and that all performance gap is not attributed to the quality of the renderings. However, it is important to stress that disentangling how much of the performance gap stems from image quality or scenarios being outside the training distribution remains challenging. 

\begin{table}[t]
    \small
    \centering
    \caption{Results for 3D object detection models fine-tuned with image-to-image translated data on laterally shifted views.}
    \setlength{\tabcolsep}{2pt} 
    \scalebox{0.8}{%
    \begin{tabular}{c c c | ccc | ccc | ccc}
    \hline 
    \toprule 
    \multirow{2}{*}{Data} & \multirow{2}{*}{\begin{tabular}{c}Shift \\ (m)\end{tabular}} & \multirow{2}{*}{FID$\downarrow$} & \multicolumn{3}{c}{FCOS3D}    & \multicolumn{3}{c}{PETR}     & \multicolumn{3}{c}{BEVFormer} \\  
                          &              &             & mAP     & NDS     & DA        & mAP     & NDS     & DA       & mAP    & NDS    & DA          \\ \midrule 
    Real                  & $0$          & -           & $30.9$  & $38.4$  & $100.0$   & $35.5$  & $40.1$  & $100.0$  & $35.6$ & $45.6$ & $100.0$   \\
    Sim                   & $0$          & $52.0$      & $23.3$  & $33.1$  & $61.6$    & $27.5$  & $35.0$  & $71.3$   & $32.3$ & $43.7$ & $77.4$    \\ 
    Sim                   & $\pm 1$      & $77.0$      & $19.9$  & $30.5$  & $42.6$    & $25.4$  & $33.3$  & $63.5$   & $29.6$ & $41.6$ & $68.3$    \\ 
    Sim                   & $\pm 2$      & $95.7$      & $16.5$  & $28.8$  & $39.7$    & $22.2$  & $31.1$  & $53.8$   & $26.9$ & $39.5$ & $59.5$    \\ 

    \bottomrule 
    \end{tabular}}
    \label{tab:laneshifted-results-3dod}
\end{table}

\begin{table}[t]
    \small
    \setlength\extrarowheight{-3pt}
    \centering
    \caption{mAP results for MapTRv2 fine-tuned with image-to-image translated data evaluated on laterally shifted camera views. The performance deteriorates as the lateral shift increases for both the original and geographically disjoint (Geogr. splits).}
    \setlength{\tabcolsep}{4pt} 
    \begin{tabular}{c c c c  c}
    \hline 
    \toprule 
    Data & Shift  (m) & FID$\downarrow$ & Original & Geogr.  \\ \midrule 
    Real & $0$     & -       & $60.6$   & $26.9$ \\ 
    Sim  & $0$     & $52.0$  & $59.2$   & $25.7$\\ 
    Sim  & $\pm 1$ & $77.0$  & $51.5$   & $21.2$\\ 
    Sim  & $\pm 2$ & $95.7$  & $39.6$   & $19.2$\\ 
    \bottomrule 
    \end{tabular}
    \vspace{-15pt}
    \label{tab:laneshifted-results-OM}
\end{table}

\begin{table}[t]
    \small
    \centering
    \caption{mAP performances for the different training methods on MapTRv2. Performance of rotated novel views can be improved substantially by injecting training data with simulated rotations.}
    \renewcommand{\arraystretch}{0.75}
    \setlength{\tabcolsep}{2pt} 
    \begin{tabular}{l c cc ccccc}
      \hline 
      \toprule 
       & \multirow{2}{*}{Finetuning} & Real & \multicolumn{6}{c}{Sim}    \\ \cmidrule(lr){3-3} \cmidrule(lr){4-9}
       & & $0^{\circ}$ & $0^{\circ}$ & $\pm 5^{\circ}$ & $\pm 15^{\circ}$ & $\pm 30^{\circ}$ & $\pm 90^{\circ}$ & $180^{\circ}$ \\ 
       \midrule  
      \multirow{2}{*}{\parbox[t]{2mm}{\multirow{2}{*}{\rotatebox[origin=c]{90}{Original}}}} 
            & FID$\downarrow$        & - & $58.4$ & $67.9$ & $83.8$ & $99.7$ & $126.3$ & $112.3$ \\ \cmidrule(lr){3-9}
            & Img2Img                & $62.5$ & $56.8$ & $53.9$ & $35.4$ & $19.3$ & $6.4$ & $19.5$ \\
            & NeRF                   & $64.9$ & $56.0$ & $53.1$ & $35.3$ & $19.3$ & $6.8$ & $19.1$ \\ 
            & NeRF+Rot               & $64.8$ & $56.1$ & $57.3$ & $51.5$ & $43.3$ & $35.6$ & $36.0$ \\  \cmidrule{2-9} 
      \multirow{3}{*}{\parbox[t]{2mm}{\multirow{2}{*}{\rotatebox[origin=c]{90}{Geogr.}}}} 
            & FID$\downarrow$        & - & $54.1$ & $68.6$ & $83.5$ & $99.7$ & $126.8$ & $113.3$ \\ \cmidrule(lr){3-9}
            & Img2Img              & $25.2$ & $24.3$ & $22.2$ & $17.4$ & $11.9$ & $5.8$  & $12.8$ \\
            & NeRF                   & $26.9$ & $24.6$ & $22.6$ & $17.3$ & $11.4$ & $5.2$  & $12.1$ \\ 
            & NeRF+Rot               & $27.5$ & $25.5$ & $24.1$ & $21.7$ & $18.4$ & $17.3$ & $16.4$\\ 
      \bottomrule 
      \end{tabular}
    \label{tab:maptrv2-rotation-finetuning}
\end{table}

\subsection{Real2sim gap correlation to image metrics}
\begin{figure}[htbp]
    \begin{subfigure}[b]{0.235\textwidth}
        \centering
        \includegraphics[width=\textwidth]{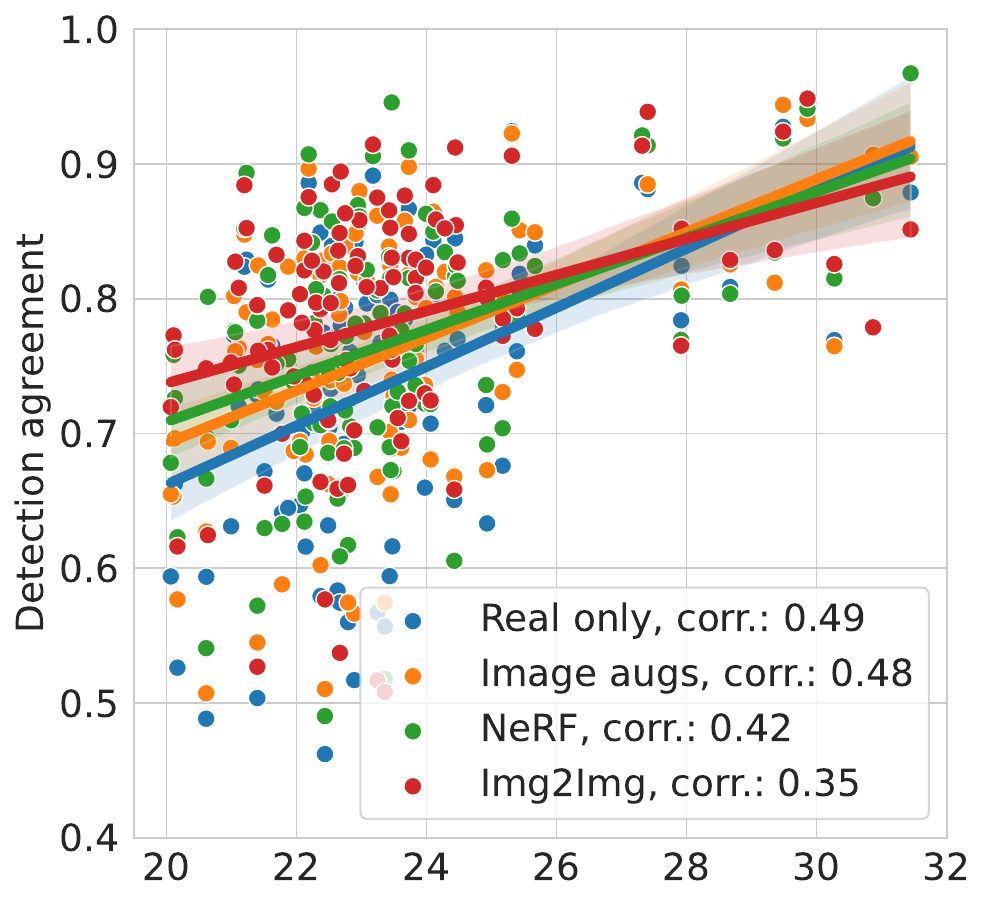}
        \caption{PSNR$\uparrow$}
        \label{fig:da-vs-psnr-bevformer}
    \end{subfigure}
    \hfill
    \begin{subfigure}[b]{0.22\textwidth}
        \centering
        \includegraphics[width=\textwidth]{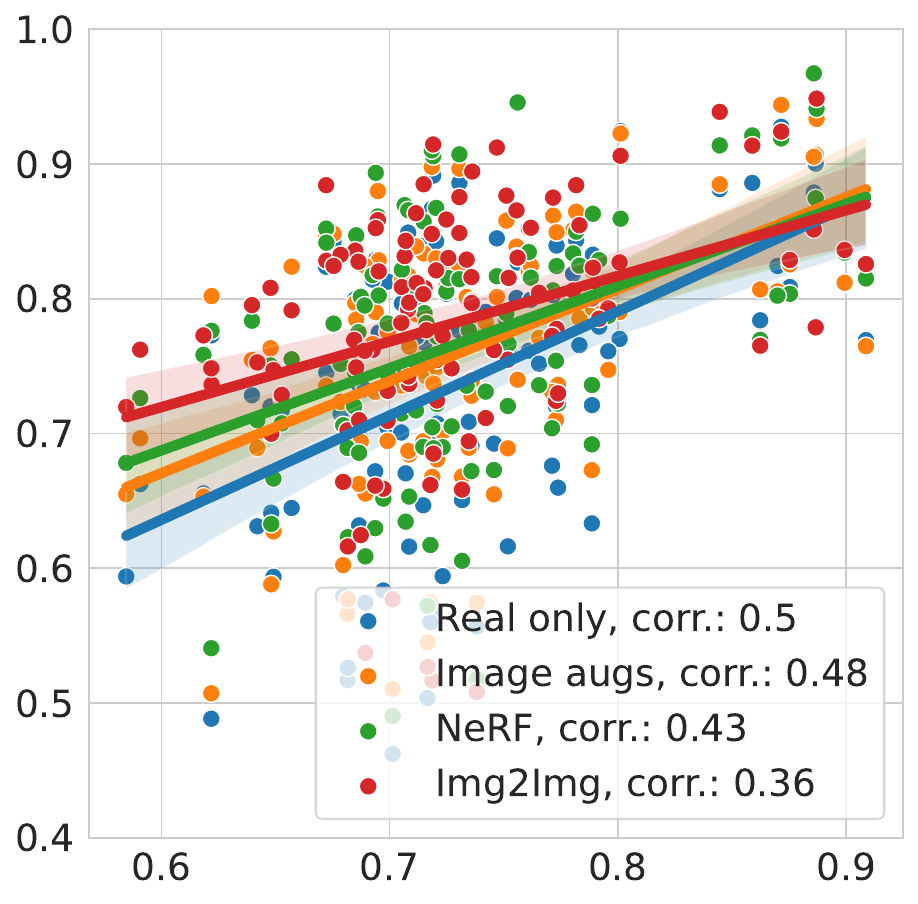}
        \caption{SSIM$\uparrow$}
        \label{fig:da-vs-ssim-bevformer}
    \end{subfigure}

    \medskip
    
    \begin{subfigure}[b]{0.23\textwidth}
        \centering
        \includegraphics[width=\textwidth]{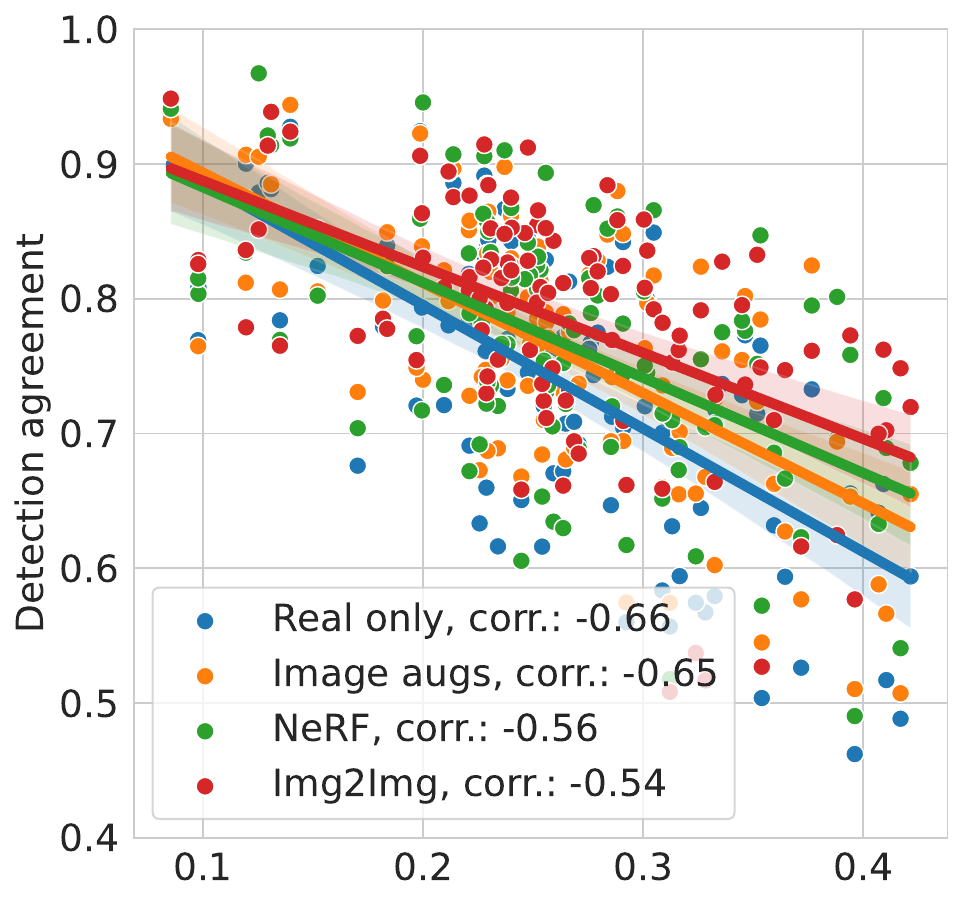}
        \caption{LPIPS$\downarrow$}
        \label{fig:da-vs-lpips-bevformer}
    \end{subfigure}
    \hfill
    \begin{subfigure}[b]{0.22\textwidth}
        \centering
        \includegraphics[width=\textwidth]{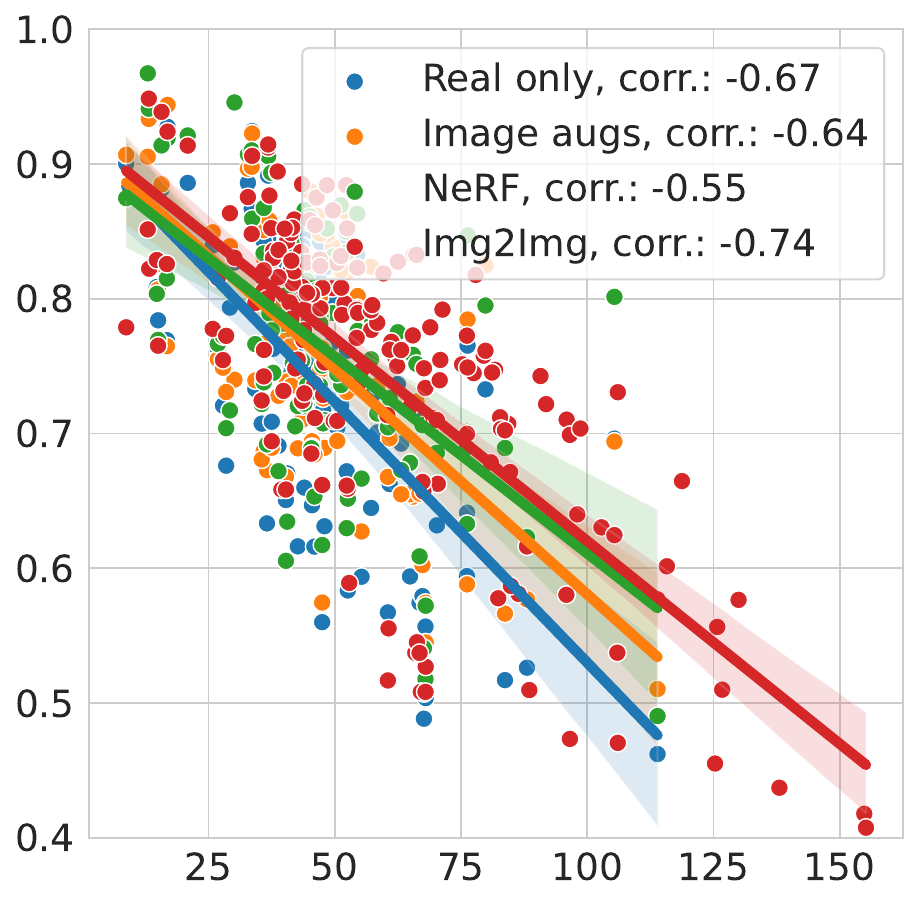}
        \caption{FID$\downarrow$}
        \label{fig:da-vs-fid-bevformer}
    \end{subfigure}
    
    \caption{Detection agreement vs. novel view synthesis metrics for BEVFormer fine-tuned with different augmentations.}
    \label{fig:da-vs-nvs-bevformer}
    \vspace{-15pt}
\end{figure}

The real2sim gap has traditionally been addressed by improving the quality of rendered images, commonly assessed using PSNR, SSIM, LPIPS and FID under the assumption that improving these metrics reduces the gap. To test this assumption, we examine the correlation between common NVS metrics and our detection agreement. Specifically, we compute PSNR, SSIM, LPIPS and FID scores for the rendered images, and detection agreement for different augmentations applied during fine-tuning of BEVFormer. Subsequently, we aggregate the results per sequence and analyze the correlation between the aggregated data points. For the model fine-tuned with our most promising method, image-to-image, we also include FID and corresponding detection agreement for the shifted scenes from \cref{sec:real2sim_extrap}, enabling us to measure the correlation for the 3DOD model and NeRF in a more practical and useful setting. Our findings, illustrated in \cref{fig:da-vs-nvs-bevformer} for each NVS metric and divided by augmentation, and in \cref{fig:da-vs-fid-shifts} in the supplementary material for the FID score on shifted scenes, reveal a clear correlation to detection agreement across all metrics. Notably, LPIPS and FID exhibit the strongest correlation and fewest outliers, indicating that perceptual similarity matters more to the perception model than mere reconstruction quality. Consequently, our results show that FID can be a useful indicator in the extrapolated setting where the other metrics are not applicable due to the lack of ground truth, \eg, to understand how large an extrapolation can be performed for a given requirement on detection agreement.  Moreover, our results indicate that, in the absence of our proposed augmentations, the model becomes considerably more sensitive to low-quality images.

\section{Conclusion}
Neural rendering has emerged as a promising avenue for simulating autonomous driving (AD) data.  
However, to be practically useful, one must understand how the behavior of an AD system on simulated data transfers to real data. 
Our large-scale investigations reveal a performance gap between perception models exposed to simulated and real images. 
We propose a new strategy to close the gap: increasing the perception models' robustness to NeRF simulated data.
We show that fine-tuning with NeRF, or NeRF-like, data substantially reduces the real2sim gap for object detection and online mapping methods with little to no performance degradation on real data. 
Moreover, for online mapping, we show that targeted generation of new scenarios can improve performance on real data. 
Nonetheless, rendering quality deteriorates rapidly when altering the ego-vehicle pose. 
Given our findings that low perceptual quality, \ie, LPIPS and FID scores, correlate strongly with a large real2sim gap, we argue that improving rendering quality in an extrapolation setting remains a key challenge for making NeRFs useful for testing and improving AD systems.


\parsection{Acknowledegments}
We thank Adam Tonderski and William Ljungbergh for valuable discussions. This work was partially supported by the Wallenberg AI, Autonomous Systems and Software Program (WASP) funded by the Knut and Alice Wallenberg Foundation. Computational resources were provided by NAISS at \href{https://www.nsc.liu.se/}{NSC Berzelius}, partially funded by the Swedish Research Council, grant agreement no. 2022-06725.
\FloatBarrier
{
    \small
    \bibliographystyle{ieeenat_fullname}
    \bibliography{main}
}

\clearpage
\setcounter{page}{1}
\maketitlesupplementary
\appendix

\section{Training details}

\subsection{Image augmentation hyperparameters}
\label{sec:img_aug_hyperparameters}
\cref{tab:img_aug_hyperparameters} shows hyperparameter selection for the image augmentation method. 

\begin{table}[t]
    \centering
    \caption{Hyperparameters for image augmentation during fine-tuning. $p$'s are referring to the probability of an augmentation being applied.}
    \resizebox{0.99\linewidth}{!}{
    \begin{tabular}{c|c}
        \toprule
        Parameter & Value  \\ 
        \midrule
        Gaussian noise $p$ & 0.5 \\ 
        Additive Gaussian noise $\delta$ & $\sim\mathcal{N}(0,10)$ \\
        \midrule
        Gaussian blur $p$ & 0.5 \\
        Gaussian blur kernel size & $5\times 5$ \\
        \midrule
        Down- and upsampling factor & $10$ \\
        Down- and upsampling method & Bilinear \\
        \midrule
        Photometric, additive brightness $\delta$ & $\sim \mathcal{U}(-32,32)$ \\
        Photometric, multiplicative contrast $\delta$ & $\sim \mathcal{U}(0.5,1.5)$ \\
        Photometric, multiplicative saturation $\delta$ & $\sim \mathcal{U}(0.5,1.5)$ \\
        Photometric, additive hue $\delta$ & $\sim \mathcal{U}(-18,18)$ \\
        \bottomrule
    \end{tabular}
    }
    \label{tab:img_aug_hyperparameters}
\end{table}

\subsection{NeRF augmentation}
\label{sec:nerf_aug_details}
To generate NeRF-rendered training images for the perception models, we train NeuRAD~\cite{tonderski2023neurad} on a subset of the nuScenes~\cite{caesar2020nuscenes} train set. We select all scenes that were collected at daytime and do not have any rain, resulting in 491 out of 750 scenes. Then we overlap this set with the geographical train split proposed in \cite{lilja2023localization}, leaving us with 316 scenes. This way, we can use the NeRF rendering both for the 3D object detection and online mapping task. Among the 316 scenes, we randomly select 110 scenes, namely \texttt{0402, 0323, 0252, 0048, 0419, 0856, 0949, 0769, 0435, 0812, 0284, 0394, 0673, 0250, 0288, 0006, 0400, 0736, 0264, 0527, 0359, 0290, 0990, 0256, 0234, 0731, 0300, 0439, 0244, 0698, 0525, 0122, 0075, 0254, 0055, 0163, 0740, 0978, 0712, 0544, 0976, 0021, 0292, 0848, 0792, 0066, 0405, 0200, 0675, 0260, 0375, 0542, 0710, 0988, 0242, 0294, 0381, 0165, 0685, 0157, 0053, 0388, 0286, 0304, 0507, 0298, 0706, 0665, 0790, 0218, 0190, 0034, 0687, 0421, 0671, 0032, 0236, 0505, 0854, 0726, 0044, 0351, 0384, 0805, 0539, 0203, 0407, 0373, 0246, 0361, 0767, 0139, 0194, 0701, 0058, 0230, 0228, 0716, 0392, 0437, 0302, 0060, 0192, 0655, 0240, 0128, 0296, 0787, 0206, 0679}. The selected sequences result in a total of 26478 images, constituting $15.7$\% of the original training set.

\subsection{Image-to-image training}
\label{sec:image_to_image_train_details}
We use the NeRF renderings outlined in \cref{sec:nerf_aug_details} to train a pix2pixHD model~\cite{wang2018high}. The 110 scenes with six cameras result in 26478 training samples for the pix2pixHD model. We use the official implementation\footnote{\url{https://github.com/NVIDIA/pix2pixHD}} and train the base model for 80 epochs, followed by tuning at a higher resolution for 45 epochs. 

\subsection{Image-to-image augmentation}
\label{sec:image_to_image_aug_details}
Using our image-to-image model, trained as outlined in \cref{sec:image_to_image_train_details}, we generate images for all 750 scenes in the nuScenes training set.

\section{Experiment details}

\subsection{Evaluation scenes}
 \label{sec:eval_scenes}
To validate the perception models, we train NeuRAD~\cite{tonderski2023neurad} on multiple nuScenes validation scenes and generate images for annotated frames. Note that these frames are held out from the NeuRAD training. From the original 150 nuScenes validation scenes, we select all scenes collected at day-time without rainy weather, yielding 111 scenes, namely \texttt{0003, 0012, 0013, 0014, 0015, 0016, 0017, 0018, 0035, 0036, 0038, 0039, 0092, 0093, 0094, 0095, 0096, 0097, 0098, 0099, 0100, 0101, 0102, 0103, 0104, 0105, 0106, 0107, 0108, 0109, 0110, 0221, 0268, 0269, 0270, 0271, 0272, 0273, 0274, 0275, 0276, 0277, 0278, 0329, 0330, 0331, 0332, 0344, 0345, 0346, 0519, 0520, 0521, 0522, 0523, 0524, 0552, 0553, 0554, 0555, 0556, 0557, 0558, 0559, 0560, 0561, 0562, 0563, 0564, 0565, 0770, 0771, 0775, 0777, 0778, 0780, 0781, 0782, 0783, 0784, 0794, 0795, 0796, 0797, 0798, 0799, 0800, 0802, 0916, 0917, 0919, 0920, 0921, 0922, 0923, 0924, 0925, 0926, 0927, 0928, 0929, 0930, 0931, 0962, 0963, 0966, 0967, 0968, 0969, 0971, 0972}. 

For the geographical split \cite{lilja2023localization}, we use all scenes in the geographical validation split that are collected at day-time without rain. Further, as NeuRAD requires annotations for training, we remove all scenes without annotations, resulting in 67 scenes. For more rigorous evaluation, we also include all scenes from the geographical test split that have annotations and were collected at day-time without rain. This adds another 87 scenes, totaling in 154 scenes, namely \texttt{0002, 0019, 0043, 0046, 0061, 0151, 0158, 0159, 0348, 0355, 0356, 0357, 0358, 0377, 0385, 0945, 0947, 0981, 0982, 0983, 0018, 0036, 0268, 0275, 0276, 0344, 0345, 0411, 0182, 0183, 0315, 0423, 0424, 0425, 0860, 0861, 0862, 0863, 0864, 0925, 0926, 0927, 0928, 0071, 0170, 0171, 0172, 0173, 0174, 0175, 0209, 0210, 0211, 0212, 0500, 0501, 0518, 0660, 0661, 0662, 0663, 0664, 0738, 0821, 0109, 0331, 0523, 0007, 0008, 0009, 0024, 0025, 0026, 0027, 0028, 0029, 0030, 0042, 0050, 0057, 0123, 0124, 0154, 0155, 0364, 0365, 0370, 0379, 0380, 0383, 0952, 0953, 0955, 0956, 0957, 0958, 0959, 0960, 0016, 0966, 0413, 0414, 0415, 0416, 0417, 0184, 0185, 0187, 0188, 0316, 0427, 0428, 0429, 0430, 0858, 0919, 0920, 0921, 0924, 0069, 0073, 0176, 0207, 0208, 0213, 0263, 0396, 0397, 0398, 0509, 0528, 0529, 0530, 0531, 0532, 0533, 0534, 0535, 0536, 0658, 0744, 0746, 0747, 0749, 0750, 0751, 0752, 0757, 0758, 0759, 0760, 0817, 0110, 0330}.

For the evaluations on laterally shifted views, we use a smaller subset of scenes from our previously chosen 111 scenes. We select scenes on the criteria that our lateral shifts do not result in the camera ending up inside other road users or structures. This results in 14 scenes, namely \texttt{0523, 0924, 0921, 0928, 0268, 0919, 0109, 0926, 0018, 0344, 0345, 0016, 0276, 0925}.

\subsection{Fine-tuning of 3D object detection models}
\label{sec:3dod_training_details}
We start all fine-tunings from model weights pre-trained on nuScenes. For FCOS3D and PETR, we utilize the implementations from the mmdetection3d-framework\footnote{\url{https://github.com/open-mmlab/mmdetection3d}} and use the model weights and corresponding training configurations reported there. For BEVFormer we use the official implementation\footnote{\url{https://github.com/fundamentalvision/BEVFormer}} and model weights corresponding to the small version. See \cref{tab:3dod_finetuning_lr} for details on the hyperparameters used for each fine-tuning.

\begin{table}[t]
    \centering
    \caption{Hyperparameters used to fine-tune the 3D object detection models.}
    \begin{tabular}{c c c c}
    \toprule
       Model                   & Augmentation & Learning rate & Epochs \\ \midrule
       \multirow{4}{*}{FCOS3D} & None         & $2e-6$        & 6 \\
                               & Image aug.  & $2e-5$        & 6 \\
                               & NeRF         & $1e-4$        & 6 \\
                               & Img2Img      & $1e-4$        & 6 \\ \midrule
       \multirow{4}{*}{PETR}   & None         & $1e-8$        & 12 \\
                               & Image aug.  & $2e-5$        & 12 \\
                               & NeRF         & $2e-5$        & 12 \\
                               & Img2Img      & $2e-5$        & 12 \\ \midrule
       \multirow{4}{*}{BEVFormer} & None      & $2e-5$        & 4 \\
                               & Image aug.  & $4e-5$        & 4 \\
                               & NeRF         & $4e-5$        & 4 \\
                               & Img2Img      & $4e-5$        & 4 \\
    \bottomrule
    \end{tabular}
    \label{tab:3dod_finetuning_lr}
\end{table}

\subsection{Fine-tuning of online mapping}
Also for the online mapping method MapTRv2 we start from the pre-trained weights for both the original and geographically disjoint splits. \cref{tab:om_finetuning_lr} reports the hyperparameters used for the different fine-tunings.

\begin{table}[t]
    \centering
    \caption{Hyperparameters used to fine-tune the online mapping method MapTRv2.}
    \begin{tabular}{c c c c}
    \toprule
       Model                   & Augmentation & Learning rate & Epochs \\ \midrule
       \multirow{4}{*}{Original} & None       & $1e-5$     & 5 \\
                               & Image aug.   & $1e-4$     & 5 \\
                               & NeRF         & $1e-4$     & 5 \\
                               & Img2Img      & $1e-4$     & 5 \\ \midrule
       \multirow{4}{*}{Geogr.}   & None       & $1e-4$     & 5 \\
                               & Image aug.   & $1e-4$     & 5 \\
                               & NeRF         & $1e-4$     & 5 \\
                               & Img2Img      & $1e-3$     & 5 \\ 
    \bottomrule
    \end{tabular}
    \label{tab:om_finetuning_lr}
\end{table}

\subsection{NeuRAD results}
In \cref{tab:neurad_results}, we report standard novel view synthesis metrics for the different data splits. We observe NeuRAD to perform similar for all data subsets, hence expecting the artifacts in the images used for augmentation having similar style as the ones used for evaluation.

\begin{table}[t]
    \caption{Novel view synthesis performance for NeuRAD on held-out images for the different splits.}
    \centering
    \begin{tabular}{c|ccc}
    \toprule
    Split & PSNR $\uparrow$ & SSIM $\uparrow$ & LPIPS $\downarrow$ \\ \midrule
     Orig. val    & 26.50  & 0.7893 & 0.2566 \\
     Orig. train & 26.52 & 0.7975 & 0.2456 \\
     Geo. val+test & 26.88 & 0.8009 & 0.2558 \\ \bottomrule
    \end{tabular}
    \label{tab:neurad_results}
\end{table}

\section{Additional results}
\subsection{Real2sim 3DOD}
\begin{figure*}
    \centering
    \includegraphics[width=0.9\linewidth]{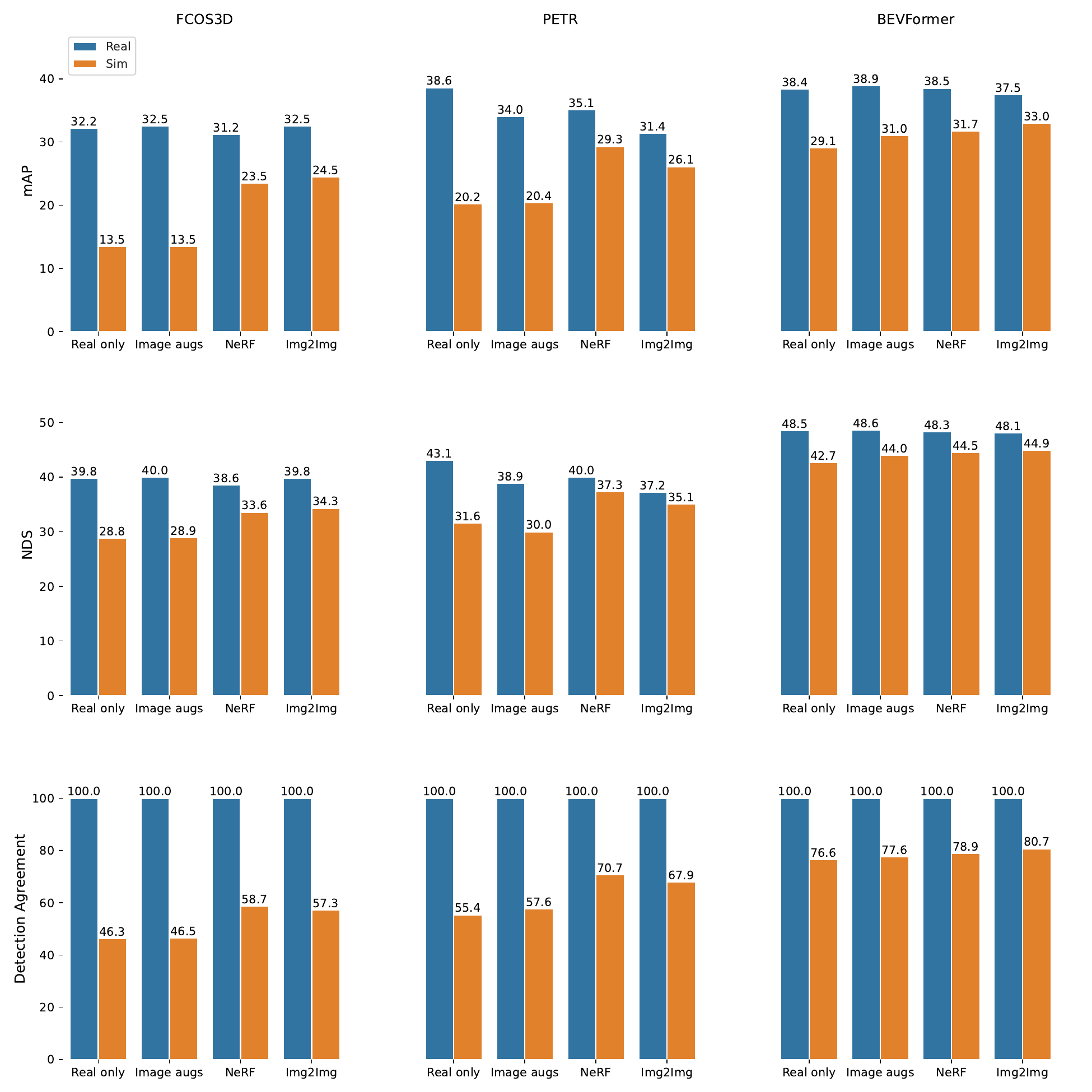}
    \caption{Bar plots of the real2sim gap for the 3DOD-models.}
    \label{fig:bar-plot-3dod}
\end{figure*}

We visualize the real2sim results on 3DOD models, reported in \cref{tab:real2sim-results-3dod}, as bar plots in \cref{fig:bar-plot-3dod}.

\subsection{Correlation to FID for shifted scenes}
\begin{figure}
    \centering
    \includegraphics[width=0.9\linewidth]{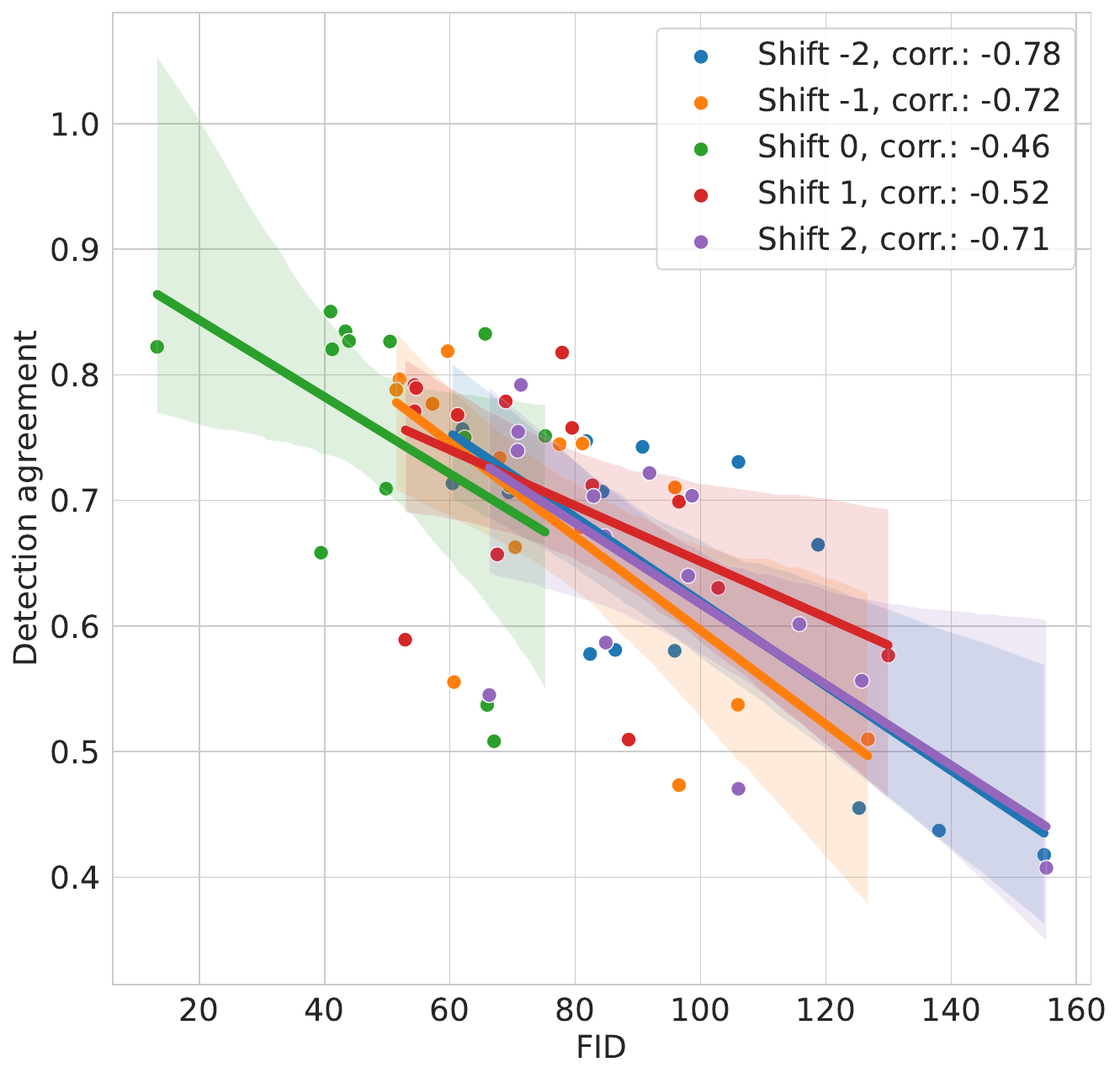}
    \caption{Detection agreement vs. FID scores for BEVFormer fine-tuned with image-to-image translated images, evaluated on the lane shift evaluation set with different shifts.}
    \label{fig:da-vs-fid-shifts}
\end{figure}

We illustrate the correlation between detection agreement and FID, isolated for only shifted sequences and divided by the shift amount and direction, in \cref{fig:da-vs-fid-shifts}. The detection agreement is computed for BEVFormer fine-tuned with image-to-image translated images.

\end{document}